\documentclass[11pt,a4paper]{article}

\usepackage[utf8]{inputenc} %
\usepackage[bottom]{footmisc}

\PassOptionsToPackage{table}{xcolor}
\PassOptionsToPackage{most}{tcolorbox}

\usepackage[nocustomfonts]{nxai}

\usepackage{url}            %
\usepackage{booktabs}       %
\usepackage{amsmath,amsfonts,bm}
\usepackage{nicefrac}       %
\usepackage{acronym}        %
\usepackage{float}
\usepackage{dblfloatfix}    %
\usepackage{multirow}
\usepackage{siunitx}        %
\usepackage{makecell}
\usepackage[colorinlistoftodos]{todonotes}
\usepackage{lineno}
\usepackage{subcaption}
\usepackage{wrapfig}
\usepackage{algorithm}
\usepackage{algpseudocode}
\usepackage{listings}
\tcbuselibrary{listings,skins,breakable}

\definecolor{algobg}{HTML}{F4F3FE}
\newtcolorbox{algobox}{
    enhanced,
    colback=algobg,
    colframe=gray!40,
    boxrule=0.4pt,
    arc=1mm,
    left=1mm,
    right=1mm,
    top=1mm,
    bottom=1mm
}

\definecolor{codebg}{RGB}{248,248,248}
\definecolor{codeframe}{RGB}{210,210,210}
\definecolor{codekw}{RGB}{0,92,175}
\definecolor{codestr}{RGB}{150,60,20}
\definecolor{codecom}{RGB}{140,140,140}

\lstdefinestyle{pytorchstyle}{
    language=Python,
    basicstyle=\ttfamily\scriptsize,
    keywordstyle=\bfseries\color{codekw},
    commentstyle=\ttfamily\color{codecom},
    stringstyle=\color{codestr},
    showstringspaces=false,
    columns=fullflexible,
    keepspaces=true,
    breaklines=true,
    breakatwhitespace=true,
    tabsize=4,
    numbers=left,
    numberstyle=\tiny\color{gray},
    numbersep=6pt,
    xleftmargin=1.5em,
    framexleftmargin=1.5em,
}

\newtcblisting{pytorchbox}{
    listing only,
    listing engine=listings,
    enhanced,
    colback=codebg,
    colframe=codeframe,
    boxrule=0.4pt,
    arc=1mm,
    left=0.5mm,
    right=0.5mm,
    top=0.5mm,
    bottom=0.5mm,
    listing options={style=pytorchstyle}
}

\acrodef{llm}[LLM]{large language model}
\acrodef{swa}[SWA]{sliding window attention}
\acrodef{rnn}[RNN]{recurrent neural network}
\acrodef{mlp}[MLP]{multi-layer perceptron}
\acrodef{rope}[RoPE]{rotary position embedding}
\acrodef{mse}[MSE]{mean-squared error}
\acrodef{ce}[CE]{cross-entropy}
\acrodef{kl}[KL]{Kullback-Leibler divergence}
\acrodef{ppl}[PPL]{perplexity}
\acrodef{peft}[PEFT]{parameter-efficient fine-tuning}
\acrodef{fft}[FFT]{full fine-tuning}
\acrodef{lora}[LoRA]{low-rank adaptation}
\acrodef{ttft}[TTFT]{time to first token}
\acrodef{oom}[OOM]{out of memory}
\acrodef{rl}[RL]{reinforcement learning}
\acrodef{ssm}[SSM]{state space model}
\acrodef{fsdp}[FSDP]{fully sharded data parallel}
\acrodef{kv}[KV]{key-value}

\usepackage{acronym}  %

\newcommand\ourmethod{\textsc{KVpop}}
\newcommand\xlstm{\textsc{xLSTM}}
\newcommand\mlstm{\textsc{mLSTM}}

\newcommand\QwenThree{\textsc{Qwen3}}
\newcommand\QwenThreeFourB{\textsc{Qwen3-4B}}
\newcommand\QwenThreeEightB{\textsc{Qwen3-8B}}

\definecolor{inferblue}{RGB}{0,92,175}

\graphicspath{{./}{figures/}}
\makeatletter
\def\input@path{{./}}
\makeatother

\newcommand{\affmark}[1]{\raisebox{0.55ex}{\fontsize{7}{7}\selectfont #1}}

\nxaisetalogo{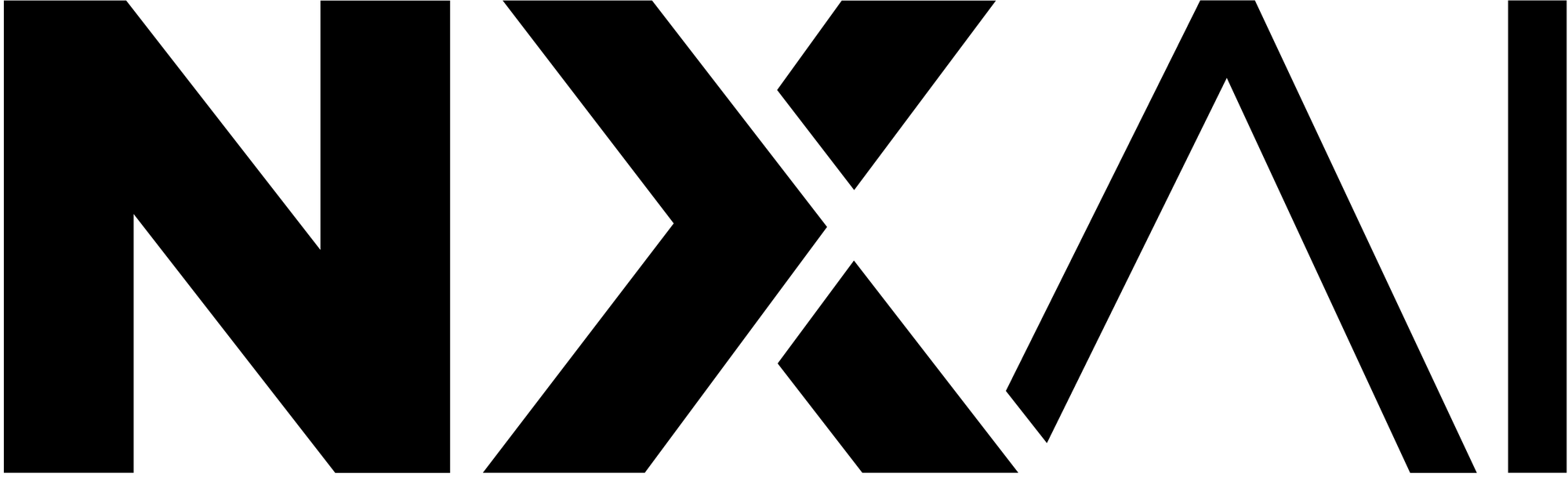}
\nxaisetlogoheight{8mm}
\nxaisetdate{July 6, 2026}
\nxaisetfooter{Correspondence: Lukas Hauzenberger (lukas.hauzenberger@nx-ai.com)}

\title{\ourmethod~--- Key-Value Cache Compression with Predictive Online Pruning}

\author{%
    Lukas Hauzenberger\affmark{1,2}\and
    Niklas Schmidinger\affmark{1,2}\and
    Anamaria-Roberta Hartl\affmark{2}\and
    David Stap\affmark{1}\and
    Thomas Schmied\affmark{$\dagger$}\and
    Sebastian Böck\affmark{1}\and
    Günter Klambauer\affmark{1,2}\and
    Sepp Hochreiter\affmark{1,2}
}
\nxaisetaffiliations{%
\affmark{1}NXAI \quad \affmark{2}Johannes Kepler University Linz, Austria \quad \affmark{$\dagger$}Work done while at NXAI
}

\nxaiabstract{%
Key-value (\acs{kv}) cache growth is a major bottleneck in autoregressive decoding, as memory and bandwidth scale linearly with context length.
Existing \acs{kv} eviction methods often rely on static heuristics or proxy scores, which poorly track future token utility and cause brittle eviction as relevance shifts.
To address this, we introduce \ourmethod, which learns a fixed-budget \acs{kv} eviction policy by directly supervising the keep-or-drop decision.
The scorer is trained against a novel future-attention target, computed efficiently without materializing dense attention maps.
We further introduce a delayed memory-based scorer that, uniquely among learned eviction methods, defers scoring for a fixed number of steps to exploit near-future context.
On AIME and HMMT mathematical reasoning, \ourmethod~retains 98\% of full-attention performance on \QwenThreeFourB{} at 75\% \acs{kv} cache compression and 97\% at 88\% compression, consistently outperforming established eviction baselines.
\QwenThreeEightB{} shows even stronger results, reaching near-full teacher performance.
These results show that supervising eviction with future-attention signals cuts memory costs while maintaining quality.
}

\nxaisethuggingface[sirluk/Qwen3-8B-KVpop-4x]{https://huggingface.co/sirluk/Qwen3-8B-KVpop-4x}

\begin{document}

\maketitle
\section{Introduction}
\label{sec:introduction}

Transformer-based \acp{llm} rely on a key--value (\acs{kv}) cache to make autoregressive decoding efficient.
At each generation step, the model stores the key and value representations of previous tokens, allowing future queries to attend to the past without recomputing the full sequence history~\citep{vaswani2017attention}.
Although this mechanism is essential for practical token-by-token generation, the \acs{kv} cache grows linearly with context length, becoming a bottleneck for long-context inference~\citep{kwon2023efficient}.
A natural way to reduce this cost is to \emph{evict} unimportant tokens and retain only a bounded subset, but token utility is hard to predict. 
Locally salient tokens may quickly become irrelevant, while tokens that receive little immediate attention may matter many steps later.
\acs{kv} cache reduction is therefore a question of which tokens will be useful for future queries.

\textbf{\acs{kv} cache reduction with supervised eviction policies.}
Existing approaches address this prediction problem in different ways.
Sliding-window and sink-token methods preserve a small set of initial tokens together with a recent window exempt from eviction~\citep{xiao2023efficient,xiao2024streaming}.
Score-based methods estimate token importance online from attention-derived or query-local signals~\citep{zhang2023h2o,oren2024tova,li2024snapkv}.
Learned approaches train retention policies during retrofitting. DMC, for example, merges token representations with learned compression policies~\citep{nawrot2024dynamic}, while DMS trains binary eviction gates and delays eviction until tokens leave a protected window of recent tokens~\citep{lancucki2025inference}.
However, none of these methods directly supervise which tokens will matter in the future. Moreover, methods with a protected window delay eviction but not the eviction decision itself. Retention of a given token is decided at insertion time, forgoing the evidence that accumulates while the token remains in the protected window.

\begin{figure*}[t]
    \centering
    \includegraphics[width=\linewidth]{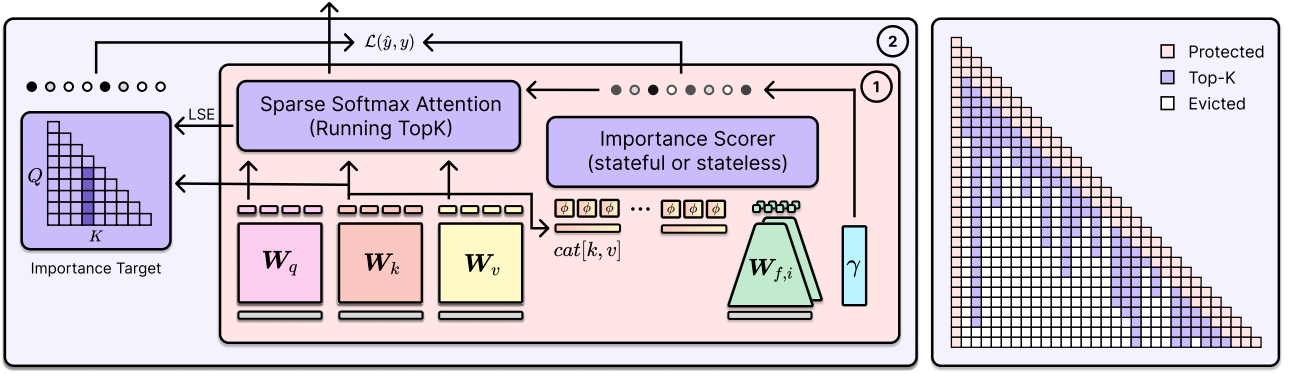}
    \caption{\textbf{Overview of \ourmethod.}
    \textbf{(1) Training and inference:} In each attention layer, keys and
    values are passed to a lightweight per-KV-head scoring module. The cache
    always retains sink tokens and a protected recent window, and uses predicted
    scores to select the remaining long-range top-$k$ entries. Attention is
    computed over the retained set.
    \textbf{(2) Training only:} Future-attention targets are computed on the fly
    and used to supervise the eviction decision induced by the fixed budget. \textbf{Sparse Attention Pattern:} \ourmethod~yields a sparse attention pattern for efficient decoding, maintaining constant memory per head during inference.}
    \label{fig:method}
\end{figure*}

\newpage
\textbf{\ourmethod.}
We introduce \ourmethod~(\textbf{KV} compression with \textbf{P}redictive \textbf{O}nline \textbf{P}runing), a sparse-attention retrofit that addresses both limitations.
Each \acs{kv} head retains a small set of sink tokens, a protected window of recent tokens, and a learned long-range top-$k$ cache for older tokens. The long-range cache is populated by lightweight head-wise scoring modules that assign importance scores to tokens and rank them under the top-$k$ budget.
This gives a bounded \acs{kv} cache for inference without changing the base architecture.
Two design choices distinguish \ourmethod~from prior learned eviction.
First, scoring module supervision is anchored to the future-attention mass a token receives after it leaves the protected window. 
The loss is evaluated at the eviction boundary where the keep-or-drop decision is made, and the target is computed during training with a transposed-attention pass that avoids materializing the dense attention map.
Second, scoring need not happen when a token enters the cache. \ourmethod~also supports stateful scorers that delay scoring until the eviction boundary, accumulating evidence while the token remains in the protected window. Unlike DMS-style gates and prior auxiliary scorers that score at insertion, eviction decisions are then informed by near-future context. An overview of \ourmethod~is shown in Figure \ref{fig:method}.

\begin{nxaiinfo}[\ourmethod~combines two ingredients]
\begin{itemize}
  \item \textbf{Cheap future-attention target.} The scoring target is computed without ever forming a dense $S\times S$ attention map. A transposed-attention pass recovers each token's post-softmax future-attention mass by reusing the LSE normalizers the attention kernel already returns.
  \item \textbf{Stateful scoring with near-future context.} \ourmethod~can optionally use a memory-based scorer that can exploit near-future context by deferring the scoring decision for a token until it reaches the eviction boundary.
\end{itemize}
\end{nxaiinfo}

\textbf{Contributions.}
We make the following contributions: \textbf{(i)} we formulate fixed-budget \acs{kv} eviction as supervised prediction at the eviction boundary, using a future-attention target and boundary-aware loss to train long-range top-$k$ decisions;
\textbf{(ii)} we compute this target with a training-only transposed-attention pass, avoiding dense attention-map materialization and inference-time overhead; 
\textbf{(iii)} we study eviction timing, showing that a delayed memory-based scorer can use near-future context before a token becomes evictable; 
\textbf{(iv)} empirically, \ourmethod~retains 95\% and 94\% of dense-attention performance on \QwenThreeFourB~at 75\% and 88\% \acs{kv} cache compression, respectively, and 95\% and 99\% on \QwenThreeEightB, outperforming heuristic and learned eviction baselines. Moreover, we show that the learned eviction policy transfers to out-of-domain code generation and STEM reasoning benchmarks while yielding nearly constant memory use and faster long-generation decoding.

\vspace{-2pt}
\section{Related Work}
\label{sec:background}

Prior work on \acs{kv} cache reduction can be broadly organized into three categories: \emph{sparse retrieval methods} reduce the number of tokens read per query while keeping the full cache available, \emph{heuristic eviction methods} reduce the persistent cache by removing tokens according to fixed rules or online scores, and \emph{learned eviction methods} train an eviction policy during retrofitting.

\textbf{Sparse retrieval over full \acs{kv} cache.}
Rather than permanently evicting tokens, sparse retrieval methods select pages, blocks, landmarks, or token groups likely to matter for the current step.
Quest retrieves relevant memory pages using lightweight approximations to token importance~\citep{tang2024quest}, and Landmark Attention introduces special tokens that route attention to relevant blocks~\citep{mohtashami2023randomaccess}.
Other approaches learn the selection rule itself: Native Sparse Attention combines coarse block selection with fine token selection in a trained sparse-attention pattern~\citep{yuan2025nsa}, DeepSeek Sparse Attention scales this idea to a deployed frontier model with a lightning indexer driving fine-grained token selection~\citep{deepseekai2025deepseekv32}. TokenButler trains a lightweight query-aware predictor of fine-grained token importance~\citep{akhauri2025tokenbutler}.
These methods reduce attention work and avoid irreversible deletion, but because the full history remains stored, they do not enforce a bounded \acs{kv} cache.

\textbf{Heuristic eviction methods.}
A second line of work imposes a bounded cache by deciding which past tokens to remove.
The simplest policies use structural priors, e.g.\ keeping recent tokens, keeping attention sinks, and evicting everything else~\citep{xiao2023efficient,xiao2024streaming}.
Other training-free methods score cached tokens online, for example using cumulative attention, current attention, or recent-query similarity, and evict low-scoring entries~\citep{zhang2023h2o,oren2024tova,li2024snapkv,cai2025pyramidkv}.
Expected Attention~\citep{devoto2025expected} estimates future attention analytically from an assumed query distribution, combined with value magnitude.
However, these scores remain proxies, since tokens that look unimportant locally may become important later, which is especially problematic in reasoning traces where earlier statements are reused after a delay.

\textbf{Learned eviction methods.}
Learned methods replace hand-designed eviction rules with policies trained during retrofitting. This improves token selection and, unlike inference-only approaches, enables the model to adapt to the train–inference mismatch introduced by \acs{kv} cache sparsification.  
Dynamic Memory Compression (DMC) learns layer- and head-specific compression policies
for pretrained models by merging unimportant tokens online~\citep{nawrot2024dynamic}.
Dynamic Memory Sparsification (DMS) evicts uninformative tokens instead of compressing them, training binary eviction gates with a differentiable relaxation and deferring removal through a sliding window so that tokens marked for eviction remain attendable for a short period~\citep{lancucki2025inference}.

\begin{nxaiinfo}[Sparse retrieval vs.\ eviction]
\begin{itemize}
    \item \textbf{Sparse retrieval} restricts each query to attend to a subset of keys, cutting attention compute and bandwidth. However, the full \acs{kv} cache stays in memory, so it imposes no memory bound.
    \item \textbf{Eviction} instead permanently drops cached tokens to enforce a hard, fixed-size cache, directly shrinking the \acs{kv} footprint. \ourmethod~is a learned eviction method.
\end{itemize}
\end{nxaiinfo}

Rather than learning eviction via differentiable relaxations, \ourmethod~trains an explicit token-level predictor of future utility, supervised by the attention mass a token receives after it exits the protected window.
The target is computed during training without materializing the dense attention map, and at decoding time \ourmethod~enforces a bounded \acs{kv} cache rather than retaining the full one.
Because the importance signal is only needed at the eviction boundary, the scorer can be delayed and integrate near-future context before the keep-or-drop decision is made.

\section{\ourmethod}
\label{sec:method}

\begin{wrapfigure}{r}{0.53\textwidth}
\vspace{-1.3em}
\hfill
\begin{minipage}{0.51\textwidth}
\centering
\begin{algobox}
\footnotesize
\input{algorithms/kvpop}
\end{algobox}
\setlength{\abovecaptionskip}{2pt}
\captionof{algorithm}{\textbf{\ourmethod~distillation.}
A scorer predicts importance scores used to construct a fixed-budget sparse attention mask. Future attention targets supervise the eviction decision. Targets are computed only during training and add no inference overhead.
}
\label{alg:kvpop}
\end{minipage}
\vspace{-1.0em}
\end{wrapfigure}

We propose \ourmethod~(\textbf{KV} compression with \textbf{P}redictive \textbf{O}nline \textbf{P}runing), a sparse-attention retrofit that enforces a fixed per-head \acs{kv} budget in pretrained language models.
Each \acs{kv} head always retains the first $s$ sink tokens and a protected window of the $w$ most recent tokens.
All remaining tokens compete for a long-range top-$k$ budget, and attention is computed only over the retained set.
Thus the per-head \acs{kv} budget during decoding is
\begin{equation}
    B = s + w + k
\label{eq:kv_budget}
\end{equation}
The retention policy is controlled by lightweight importance scorers that produce per-head rankings of eligible tokens online.
At training time, we additionally compute a future-attention supervision target that the scorers regress against.
At inference, only the budget $B$ is enforced.
Figure~\ref{fig:method} provides an overview, and Algorithm~\ref{alg:kvpop} details one training step of \ourmethod.

\subsection{Importance Scorer Supervision}
\label{sec:target_loss}

The scoring target should reflect future utility, not only past attention or local token statistics. 
We therefore define supervision in terms of the future attention mass a key receives after it has left the protected window. 
The target is independent of the scorer architecture and can supervise both stateless and stateful policies (see Section~\ref{sec:scorers}).

\begin{wrapfigure}{r}{0.58\textwidth}
    \centering
    \includegraphics[width=0.56\textwidth]{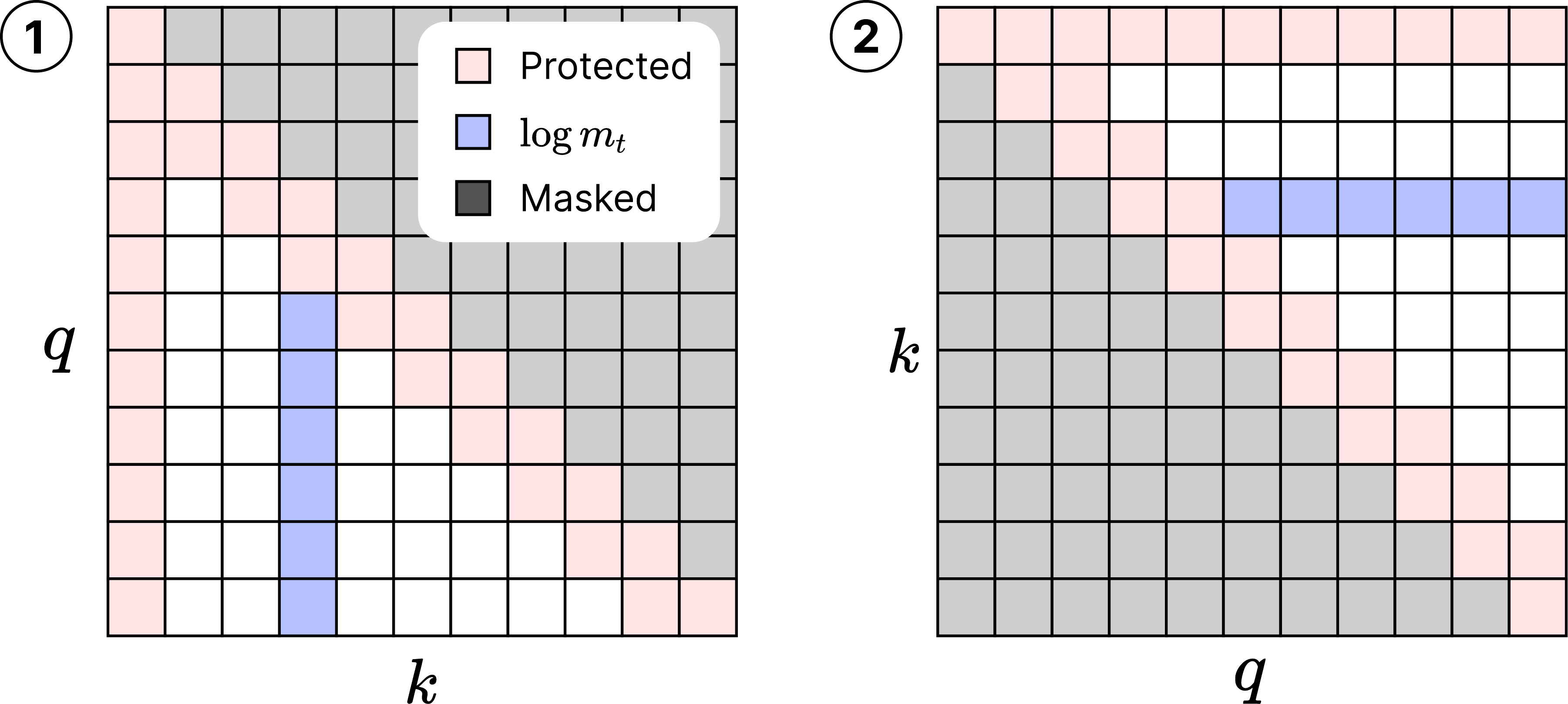}
    \caption{\textbf{Transposed Attention.}
    \textbf{(1)} Blue tiles show the future attention mass for token $t$.
    \textbf{(2)} Transposing $q$ and $k$ turns the per-key column-sum into a per-query row-sum, which attention kernels return as their auxiliary LSE.}
    \label{fig:kvpop_target}
    \vspace{-3.0em}
\end{wrapfigure}

\textbf{Future-attention target.}
Let $S$ be the training sequence length and let $g \in \{1,\ldots,G\}$ index the query heads that share \acs{kv} head $h$ under grouped-query attention \citep{ainslie2023gqa}. 
Let $p^{(h,g)}_{d\to t}$ denote the dense causal attention probability from query head $(h,g)$ at position $d$ to key $(h,t)$. Since token $t$ is protected until queries are at least $w$ positions ahead, we define its mean future-attention mass per group as
\vspace{-1pt}
\begin{equation}
\begin{aligned}
    m^{(h,g)}_t
    &=
    \frac{1}{N_t}
    \sum_{d=t+w}^{S-1}
    p^{(h,g)}_{d\to t}
    \\
    \text{where } N_t
    &=
    \max(1,S-(t+w))
\end{aligned}
\label{eq:future_mass_per_group}
\end{equation}

Since a \acs{kv} entry is shared by all $G$ query heads, we aggregate the per-group log-masses to obtain the future-utility target.
\begin{equation}
    r^{\mathrm{tgt}}_{h,t}
    =
    \operatorname{Agg}_{g}
    \left[
    \log\left(\epsilon + m^{(h,g)}_t\right)
    \right]
\label{eq:future_importance_target}
\end{equation}
where $\epsilon>0$ is a small constant. In the experiments, $\operatorname{Agg}_{g}$ is max aggregation. Alternatives are discussed in Appendix~\ref{app:target_loss_details}.

\textbf{Effective scores and teacher policy.}
At query position $q$, only tokens outside the sink region and outside the
protected recent window compete for the long-range budget:
\begin{equation}
    \mathcal{E}(q) = \{t \mid s \le t \le q-w\}
\label{eq:eligible_set}
\end{equation}
The token $t_{\mathrm{new}} = q-w$ has just left the protected window and becomes newly eligible. The teacher ranks eligible tokens by target effective scores
\begin{equation}
    r^{\mathrm{tgt}}_{h,t}(q)
    =
    r^{\mathrm{tgt}}_{h,t}
    +
    \left\lfloor \frac{q-t}{n} \right\rfloor \log \gamma_h
\label{eq:target_effective_score}
\end{equation}

where $\gamma_h \in (0,1)$ is a per-head decay factor and $n$ is the decay step size. The decay term introduces a recency bias that prevents tokens with very high scores from occupying cache slots indefinitely, allowing newer tokens to compete for retention over time.
The teacher and student share the same $\gamma_h$, ensuring the supervision matches the inference-time ranking rule. The teacher retains the top-$k$ tokens in $\mathcal{E}(q)$ under Eq.~\eqref{eq:target_effective_score}.

Let $t_{\mathrm{bnd}}$ be the boundary token at the teacher cutoff, i.e. the last retained token under the teacher ranking. The teacher label for the newly eligible token is
\begin{equation}
y_{q,h}
=
\begin{cases}
+1, & \text{if } t_{\mathrm{new}} \text{ is retained}\\
-1, & \text{otherwise}
\end{cases}
\label{eq:teacher_label}
\end{equation}

\textbf{Boundary-aware retention loss.}
Let $\hat r_{h,t}(q)$ be the predicted effective score used by the sparse attention policy, defined analogously to Eq.~\eqref{eq:target_effective_score} from the predicted raw score $\hat r_{h,t}$. 
Since the decay term is identical for any two tokens at the same query position $q$, the comparison reduces to the difference of raw scores. We train the scorer with a pairwise logistic loss at the retention boundary:
\begin{equation}
    \mathcal{L}_{\mathrm{score}}
    =
    \mathbb{E}_{q,h}
    \left[
    \omega_{q,h}
    \operatorname{softplus}\left(
    -y_{q,h}
    \frac{
    \hat r_{h,t_{\mathrm{new}}}(q)
    -
    \hat r_{h,t_{\mathrm{bnd}}}(q)
    }{\tau}
    \right)
    \right]
\label{eq:boundary_loss}
\end{equation}
where $\tau$ is a temperature and $\omega_{q,h}$ is an optional weighting term.
Setting $\omega_{q,h}=1$ gives the unweighted objective.
In our implementation, we use $\omega_{q,h}$ to downweight ambiguous teacher decisions and balance keep/drop decisions across heads.
Details are given in Appendix~\ref{app:target_loss_details}.
The boundary loss focuses capacity on the single comparison that changes cache membership and costs $O(1)$ per sampled query position once the teacher cutoff is known.

\begin{nxaiinfo}[Boundary loss intuition]
The loss looks at a single pair per query: the newly evictable token $t_{\mathrm{new}}$ and the teacher's cutoff token $t_{\mathrm{bnd}}$ (i.e. the lowest-scoring token still within the top-$k$). It pushes the student's score for $t_{\mathrm{new}}$ above the cutoff when the teacher keeps it and below when the teacher drops it. The $\operatorname{softplus}$ acts as a smooth hinge: $\approx0$ once the decision is correct by a margin, growing with the score gap when it is wrong.
\end{nxaiinfo}

\subsection{Efficient Target Implementation}
\label{sec:efficient_target}

The target in Eq.~\eqref{eq:future_importance_target} is defined through dense causal attention probabilities, but we never materialize the corresponding $S \times S$ probability matrices.
The student forward pass already computes a fixed-budget sparse attention pattern and returns per-query log-normalizers from the sparse attention kernel.
We reuse these sparse log-normalizers as an approximation to the dense causal normalizers, and compute the future-attention target with one additional transposed-attention pass implemented with efficient attention kernels such as \textsc{FlexAttention}~\citep{dong2025flexattention}.

\textbf{Transposed-attention target computation.}
Let $\ell^{(h,g)}(d,t) = \langle \boldsymbol{q}^{(h,g)}_d, \boldsymbol{k}^{(h)}_t \rangle / \sqrt{d_k}$ be the pre-softmax attention logit, and let $\mathrm{LSE}^{(h,g)}_d$ be the causal log-sum-exp (LSE) normalizer for query position $d$. The mean future-attention mass $m^{(h,g)}_t$ from Eq.~\eqref{eq:future_mass_per_group} can equivalently be written as
\begin{equation}
    \log m^{(h,g)}_t
    =
    \log \sum_{d=t+w}^{S-1}
    \exp\left(\ell^{(h,g)}(d,t) - \mathrm{LSE}^{(h,g)}_d\right)
    - \log N_t
\label{eq:future_mass_lse}
\end{equation}
The first term is a log-sum-exp over future query positions for a fixed key $t$, corresponding to a column-wise reduction over the dense logit matrix. We evaluate it by swapping the roles of queries and keys in a second attention-like call (Figure \ref{fig:kvpop_target}). Original key positions $t$ become queries, original query positions $d$ become keys, and the dot products recover the original logits $\ell^{(h,g)}(d,t)$. We subtract the per-query normalizer as a score modifier and apply a block mask enforcing $d \ge t+w$. The auxiliary LSE returned by the attention kernel then gives the first term of Eq.~\eqref{eq:future_mass_lse} for all keys in parallel, after which we subtract $\log N_t$ to obtain $\log m^{(h,g)}_t$. Adding $\epsilon$ inside the log and aggregating across query heads sharing \acs{kv} head $h$ yields $r^{\mathrm{tgt}}_{h,t}$ as in Eq.~\eqref{eq:future_importance_target}.
Using the dense causal $\mathrm{LSE}^{(h,g)}_d$ would make this identity exact.
Instead, we reuse the sparse query LSE values
\begin{equation}
\widetilde{\mathrm{LSE}}^{(h,g)}_d
=
\log \sum_{t'=0}^{d}
M_{d,t'}\exp\left(\ell^{(h,g)}(d,t')\right)
\label{eq:sparse_lse_main}
\end{equation}
already returned by the student attention pass, where $M$ is the fixed-budget sparse mask. The transposed pass therefore only requires a masked LSE through the same kernel interface, adding no inference-time overhead. We find empirically that this sparse-LSE approximation matches the dense-LSE target in downstream performance, while avoiding the cost of a separate dense-attention pass during training.
The per-group targets are aggregated across query heads sharing \acs{kv} head $h$ to produce $r^{\mathrm{tgt}}_{h,t}$ as in Eq.~\eqref{eq:future_importance_target}. Appendix~\ref{app:efficient_target} provides the full derivation.

\textbf{Running top-$k$ sparse attention.}
At each query position $q$, the sparse attention mask retains the union of the sink tokens, the protected recent window, and the top-$k$ tokens from $\mathcal{E}(q)$ under predicted effective scores. The predicted effective score has the same form as the teacher score.
For a fixed query, the query-dependent part is shared across eligible tokens up to the decay-step discretization. The top-$k$ selection therefore reduces to a running threshold over static token priorities rather than recomputing a full top-$k$ from scratch at every query.

Concretely, we sort tokens once by their static priority and compute a rank for each token. As $q$ increases, exactly one new token, $t=q-w$, enters the eligible set. We insert its rank into a Fenwick tree~\citep{fenwick1994} and retrieve the rank of the $k$-th best eligible token by binary lifting. This gives a query-specific cutoff rank $\tau_q$, so the retained long-range keys are exactly those eligible tokens with rank at most $\tau_q$. The resulting predicate is evaluated inside \textsc{FlexAttention}, which generates the sparse mask on the fly inside the fused attention kernel rather than storing a dense $S \times S$ mask. The cutoff computation costs $O(S\log S)$ time and $O(S)$ space per head, while the attention computation itself is performed by the sparse kernel. Appendix~\ref{app:kvpop_pseudocode} contains the full algorithm.

\begin{nxaiinfo}[Efficient kernel for running top-$k$]
We compute the per-query cutoff rank with an online Fenwick tree over static ranks, implemented as a custom Triton kernel parallelized over batch elements and \acs{kv} heads. The result is handed to FlexAttention, which builds the sparse mask on the fly, never storing a dense $S\times S$ matrix.
\end{nxaiinfo}

\subsection{Online Importance Scorers}
\label{sec:scorers}

\begin{figure*}[t]
    \centering
    \includegraphics[width=.95\linewidth]{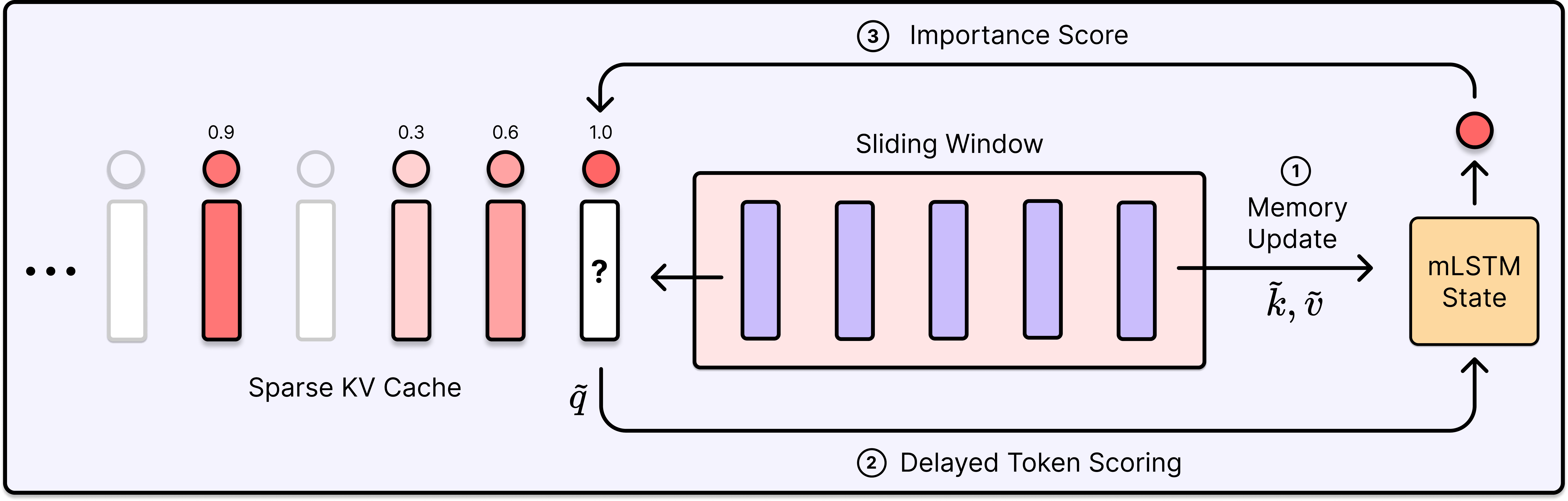}
    \caption{\textbf{\ourmethod~stateful eviction policy.}
    \textbf{(1)} The \mlstm-memory is updated with the most recent \acs{kv}
    pair.
    \textbf{(2)} For delayed scoring, the memory is read with the token at
    position $q-w$ that has just exited the protected sliding window.
    \textbf{(3)} The \mlstm~emits an importance score, and ranking determines
    whether the token is kept or evicted.}
    \label{fig:delayed_scoring}
\end{figure*}

The target and loss above can supervise any module that assigns scalar importance scores to cached tokens. We use lightweight scorers per \acs{kv} head whose inputs are derived from cached keys and values.
Crucially, the scorer inputs are detached from the computation graph, so the scorer is optimized independently of the backbone: the retention loss in Eq.~\eqref{eq:boundary_loss} updates only the scorer parameters, while the base model is trained solely by the distillation objective.

\textbf{Stateless scorers.}
The simplest scorer predicts token importance from the token's own representation. We concatenate the key and value at position $t$ as $\boldsymbol{x}_{h,t} = [\boldsymbol{k}_{h,t};\boldsymbol{v}_{h,t}]$ and compute $\hat r_{h,t}=f^{(h)}_\theta(\boldsymbol{x}_{h,t})$, where $f^{(h)}_\theta$ is a small headwise scorer. In our experiments we use a two-layer MLP with a SiLU activation. Stateless scorers are cheap, but they score each token using only local information.

\textbf{Stateful scorers.}
Stateless scorers assign an importance score from the token's representation at insertion time.
The token's own representation, however, may not capture how it relates to earlier context, and it cannot incorporate context that accumulates afterward.
A stateful scorer instead maintains a memory shaped by the retention objective, and the protected window enables \emph{delayed} scoring.
A token does not need a score when it enters the \acs{kv} cache, only when it leaves the window and begins competing for the long-range budget.
At query position $q$, the scorer can therefore update its memory with tokens up to $q$ before scoring $t_{\mathrm{new}}=q-w$, using near-future context relative to the token being scored (Figure~\ref{fig:delayed_scoring}).
This is distinct from delayed eviction as in DMS~\citep{lancucki2025inference}, which postpones the decision without incorporating the accumulated evidence.
We instantiate this with an \mlstm~\citep{beck2024xlstm} scorer for each \acs{kv} head. Let $\boldsymbol{C}_{h,q}$ and $\boldsymbol{z}_{h,q}$ denote the \mlstm~state and normalizer after processing tokens up to position $q$, and let $\tilde{\boldsymbol q}_{h,t_{\mathrm{new}}}$ be a projected feature vector of the newly eligible token.
The delayed readout is
\begin{equation}
    \boldsymbol{h}_{h,t_{\mathrm{new}}}
    =
    \frac{
    \tilde{\boldsymbol q}_{h,t_{\mathrm{new}}}^{\top}\boldsymbol{C}_{h,q}
    }{
    \tilde{\boldsymbol q}_{h,t_{\mathrm{new}}}^{\top}\boldsymbol{z}_{h,q}
    }
\label{eq:delayed_mlstm_readout}
\end{equation}
and the raw importance score is produced by a small headwise projection
\begin{equation}
\hat r_{h,t_{\mathrm{new}}}
=
\boldsymbol{a}_h^{\top}
\operatorname{SiLU}(\boldsymbol{h}_{h,t_{\mathrm{new}}})
+
b_h
\label{eq:mlstm_score}
\end{equation}
Appendix~\ref{app:scorer_architecture} gives the full \mlstm~formulation, feature maps, gates, initialization, and implementation variants. Appendix~\ref{app:xlstm} provides background on \mlstm~and Linear Attention.

\section{Experiments}
\label{sec:experiments}

\begin{table}[t]
\centering
\small
\caption{\textbf{Pass@1} on \textbf{AIME} and \textbf{HMMT} shown as absolute scores (Abs) and w.r.t. teachers (Rel)}
\label{tab:results_qwen3_math_rel}
\setlength{\tabcolsep}{3pt}
\begin{tabular}{
@{} c l
@{\hspace{6pt}\vrule\hspace{6pt}}
S[table-format=1.2] S[table-format=1.2]
@{\hspace{6pt}\vrule\hspace{6pt}}
S[table-format=1.2] S[table-format=1.2]
@{\hspace{6pt}\vrule\hspace{6pt}}
S[table-format=1.2] S[table-format=1.2]
@{\hspace{8pt}\vrule\hspace{8pt}}
S[table-format=1.2] S[table-format=1.2]
@{\hspace{6pt}\vrule\hspace{6pt}}
S[table-format=1.2] S[table-format=1.2]
@{\hspace{6pt}\vrule\hspace{6pt}}
S[table-format=1.2] S[table-format=1.2] @{}
}
\toprule
\multicolumn{2}{c}{}
& \multicolumn{6}{c}{\QwenThree{} 4B}
& \multicolumn{6}{c}{\QwenThree{} 8B} \\
\cmidrule(l{-3pt}r{10.5pt}){3-8}
\cmidrule(l{-5.5pt}r{3pt}){9-14}
& \multirow{2}{*}{Variant}
& \multicolumn{2}{c@{\hspace{6pt}\vrule\hspace{6pt}}}{AIME} & \multicolumn{2}{c@{\hspace{6pt}\vrule\hspace{6pt}}}{HMMT} & \multicolumn{2}{c@{\hspace{8pt}\vrule\hspace{8pt}}}{Average}
& \multicolumn{2}{c@{\hspace{6pt}\vrule\hspace{6pt}}}{AIME} & \multicolumn{2}{c@{\hspace{6pt}\vrule\hspace{6pt}}}{HMMT} & \multicolumn{2}{c}{Average} \\
&
& {2024} & {2025} & {2502} & {2511} & {Abs.} & {Rel.}
& {2024} & {2025} & {2502} & {2511} & {Abs.} & {Rel.} \\
\midrule
\multicolumn{1}{c}{} & Teacher
& \num{0.61} & \num{0.46} & \num{0.30} & \num{0.43} & \num{0.45} & \num{1.00}
& \num{0.58} & \num{0.49} & \num{0.28} & \num{0.37} & \num{0.43} & \num{1.00} \\
\cmidrule(lr){1-14}
\multirow{6}{*}{\rotatebox{90}{$\text{CR}=75\%$}}
& StreamLLM
& \num{0.47} & \num{0.33} & \num{0.24} & \num{0.33} & \num{0.34} & \num{0.76}
& \num{0.13} & \num{0.11} & \num{0.03} & \num{0.05} & \num{0.08} & \num{0.19} \\
& TOVA
& {\underline{\num{0.56}}} & \num{0.38} & \num{0.28} & \num{0.10} & \num{0.33} & \num{0.73}
& \num{0.37} & \num{0.23} & \num{0.15} & \num{0.28} & \num{0.26} & \num{0.60} \\
& StreamLLM+
& \num{0.55} & \bfseries\num{0.44} & {\underline{\num{0.30}}} & \num{0.36} & \num{0.41} & \num{0.92}
& \num{0.53} & \num{0.41} & \num{0.25} & {\underline{\num{0.35}}} & \num{0.39} & \num{0.91} \\
& DMS
& \bfseries\num{0.62} & \bfseries\num{0.44} & \num{0.28} & \num{0.38} & {\underline{\num{0.43}}} & {\underline{\num{0.96}}}
& \num{0.56} & \num{0.45} & \num{0.27} & {\underline{\num{0.35}}} & \num{0.41} & {\underline{\num{0.95}}} \\
& \ourmethod\textsubscript{mlp}
& \bfseries\num{0.62} & {\underline{\num{0.43}}} & \num{0.29} & \bfseries\num{0.40} & \bfseries\num{0.44} & \bfseries\num{0.98}
& \bfseries\num{0.59} & {\underline{\num{0.46}}} & {\underline{\num{0.30}}} & \bfseries\num{0.38} & {\underline{\num{0.43}}} & \bfseries\num{1.00} \\
& \ourmethod
& \bfseries\num{0.62} & \bfseries\num{0.44} & \bfseries\num{0.31} & {\underline{\num{0.39}}} & \bfseries\num{0.44} & \bfseries\num{0.98}
& {\underline{\num{0.57}}} & \bfseries\num{0.48} & \bfseries\num{0.31} & \bfseries\num{0.38} & \bfseries\num{0.44} & \bfseries\num{1.00} \\
\cmidrule(lr){1-14}
\multirow{6}{*}{\rotatebox{90}{$\text{CR}=88\%$}}
& StreamLLM
& \num{0.30} & \num{0.23} & \num{0.15} & \num{0.17} & \num{0.21} & \num{0.47}
& \num{0.13} & \num{0.11} & \num{0.03} & \num{0.05} & \num{0.08} & \num{0.19} \\
& TOVA
& \num{0.36} & \num{0.23} & \num{0.19} & \num{0.28} & \num{0.26} & \num{0.58}
& \num{0.08} & \num{0.07} & \num{0.09} & \num{0.08} & \num{0.08} & \num{0.19} \\
& StreamLLM+
& \num{0.45} & \num{0.33} & \num{0.26} & \num{0.29} & \num{0.33} & \num{0.74}
& \num{0.42} & \num{0.30} & \num{0.18} & \num{0.24} & \num{0.29} & \num{0.67} \\
& DMS
& \num{0.58} & {\underline{\num{0.41}}} & {\underline{\num{0.27}}} & \num{0.35} & \num{0.40} & \num{0.89}
& \num{0.52} & \num{0.38} & {\underline{\num{0.22}}} & \num{0.31} & \num{0.36} & \num{0.84} \\
& \ourmethod\textsubscript{mlp}
& {\underline{\num{0.59}}} & \bfseries\num{0.44} & {\underline{\num{0.27}}} & {\underline{\num{0.38}}} & {\underline{\num{0.42}}} & {\underline{\num{0.93}}}
& {\underline{\num{0.57}}} & {\underline{\num{0.39}}} & \bfseries\num{0.31} & \bfseries\num{0.40} & {\underline{\num{0.42}}} & {\underline{\num{0.98}}} \\
& \ourmethod
& \bfseries\num{0.61} & \bfseries\num{0.44} & \bfseries\num{0.30} & \bfseries\num{0.39} & \bfseries\num{0.44} & \bfseries\num{0.97}
& \bfseries\num{0.58} & \bfseries\num{0.44} & \bfseries\num{0.31} & {\underline{\num{0.39}}} & \bfseries\num{0.43} & \bfseries\num{1.00} \\
\bottomrule
\end{tabular}

\end{table}

We apply \ourmethod~distillation to \QwenThreeFourB\textsc{-Instruct-2507} and \QwenThreeEightB~\citep{yang2025qwen3technicalreport}.
We use a training sequence length of $S=16384$.
For sparsification, we use the Nemotron-Math v2 dataset~\citep{du2025nemotronmath}, selecting the subset with high reasoning effort and filtering for sequences of length at most $S$ under the Qwen tokenizer.
We apply sequence packing to improve token utilization.

We train both a stateless variant, \ourmethod\textsubscript{mlp}, and a stateful variant, \ourmethod, for 2{,}000 steps.
Since recurrent state introduces additional memory overhead for stateful scorers, we reduce the top-$k$ budget of the stateful variants to match the memory footprint of the stateless scorers.
We use a cosine schedule for the scorer parameters with peak learning rate $10^{-3}$, keep the base-model learning rate fixed at $8\times10^{-5}$, and include a KL-divergence loss.
Full hyperparameters are provided in Appendix~\ref{app:experiment_details}.

\textbf{Baselines.}
We compare against both training-free and trained \acs{kv} cache baselines and the original full-attention teachers.
Sparse-retrieval methods are excluded from comparison since they retain the full \acs{kv} cache.
For \QwenThreeEightB, which is a hybrid reasoning model, we enable thinking mode. As
training-free eviction methods, we use StreamingLLM~\citep{xiao2023efficient},
which keeps attention sinks and a recent sliding window, and
TOVA~\citep{oren2024tova}, which evicts tokens according to attention
scores once the cache budget is reached. To isolate the effect of learned scoring from training under a fixed sparse pattern, we additionally train a StreamingLLM variant (StreamingLLM+) with the same budget and setup as \ourmethod. Finally, we compare against Dynamic Memory Sparsification (DMS)~\citep{lancucki2025inference}, a learned \acs{kv} cache sparsification method that uses differentiable relaxations to train the eviction policy. We train DMS with the same budget as \ourmethod~and match its parameter count to the \ourmethod~scorers, so differences reflect the retention objective rather than scorer capacity.

\subsection{Results}

\textbf{Mathematical Reasoning.}
We report \textbf{pass@1} on AIME24/25 and HMMT (Feb/Nov 2025) in Table~\ref{tab:results_qwen3_math_rel} under two cache budgets $B\in\{2048,4096\}$, corresponding to compression ratios (CR) of 88\% and 75\%. Pass@1 is estimated using 16 rollouts per prompt. We report absolute and relative scores w.r.t. teachers.

At CR=88\%, training-free StreamingLLM and TOVA collapse, and StreamingLLM+ recovers only partially.
\ourmethod~preserves 97\% and 100\% of \QwenThreeFourB~and \QwenThreeEightB~respectively.
The stateless \ourmethod\textsubscript{mlp} reaches 93\% and 98\%. The progression from StreamingLLM to StreamingLLM+ to \ourmethod~isolates the contribution of training.
While training the model with a fixed attention pattern recovers some of the gap, a learned eviction policy effectively closes it.
Moreover, the results suggest that the supervision target of the eviction policy also matters.
DMS trains its policy via a Gumbel-sigmoid instead of distilling from teacher attention and trails \ourmethod~by 8-16 points in relative performance.

At CR=75\%, all methods improve, but the ordering holds, \ourmethod~outperforms all sparse baselines on average.

\begin{wraptable}{r}{0.41\textwidth}
\vspace{-2pt}
\centering
\caption{Results for \QwenThreeFourB~on \textbf{GPQA-D} and \textbf{LCB}.}
\label{tab:results_qwen3_other}
\setlength{\tabcolsep}{3pt}
\begin{tabular}{
@{} c l
S[table-format=1.2] S[table-format=1.2] @{}
}
\toprule
& Variant
& {GPQA-D} & {LCB} \\
\midrule
\multicolumn{1}{c}{} & Teacher
& \num{0.59} & \num{0.35} \\
\cmidrule(lr){1-4}
\multirow{5}{*}{\rotatebox{90}{$\text{CR}=75\%$}}
& StreamLLM
& \num{0.54} & \underline{\num{0.36}} \\
& TOVA
& \underline{\num{0.58}} & \num{0.34} \\
& StreamLLM+
& \num{0.55} & \num{0.32} \\
& DMS
& \num{0.55} & \bfseries \num{0.37} \\
& \ourmethod\textsubscript{mlp}
& \bfseries \num{0.59} & \num{0.35} \\
& \ourmethod
& \num{0.57} & \num{0.33} \\
\cmidrule(lr){1-4}
\multirow{5}{*}{\rotatebox{90}{$\text{CR}=88\%$}}
& StreamLLM
& \num{0.49} & \bfseries \num{0.35} \\
& TOVA
& \num{0.54} & \bfseries \num{0.35} \\
& StreamLLM+
& \num{0.52} & \num{0.29} \\
& DMS
& \num{0.54} & \bfseries \num{0.35} \\
& \ourmethod\textsubscript{mlp}
& \bfseries \num{0.57} & \bfseries \num{0.35} \\
& \ourmethod
& \underline{\num{0.56}} & \underline{\num{0.34}} \\
\bottomrule
\end{tabular}

\vspace{-14pt}
\end{wraptable}

\textbf{Sparsification generalization beyond mathematical reasoning.}
To test whether the learned sparsification policy generalizes beyond the training data distribution and the main mathematical reasoning benchmarks, we also evaluate the sparsified \QwenThreeFourB~models on out-of-domain reasoning tasks.
Following the general-purpose evaluation protocol of DMS~\citep{lancucki2025inference}, we report accuracy on GPQA Diamond (GPQA-D)~\citep{rein2024gpqa} and pass@1 on LiveCodeBench v6 (LCB)~\citep{jain2025livecodebench} in Table~\ref{tab:results_qwen3_other}.
These benchmarks probe scientific reasoning and code generation, respectively, and therefore cover domains not included in the sparsification training data.
Across both benchmarks and compression ratios, \ourmethod~stays close to the dense teacher despite being trained only on mathematical reasoning data.
Eviction methods perform comparably on these tasks, which require smaller reasoning budgets than AIME and HMMT.

\begin{figure*}
    \centering

    \begin{subfigure}[t]{0.48\linewidth}
        \centering
        \includegraphics[width=\linewidth]{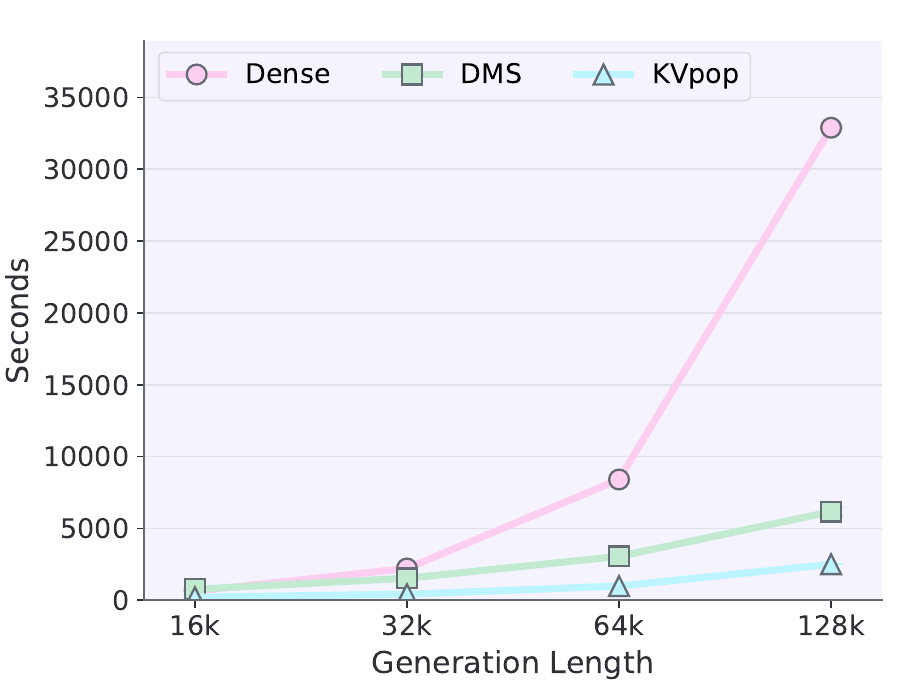}
        \caption{End-to-end latency.}
        \label{fig:inference_latency}
    \end{subfigure}
    \hfill
    \begin{subfigure}[t]{0.48\linewidth}
        \centering
        \includegraphics[width=\linewidth]{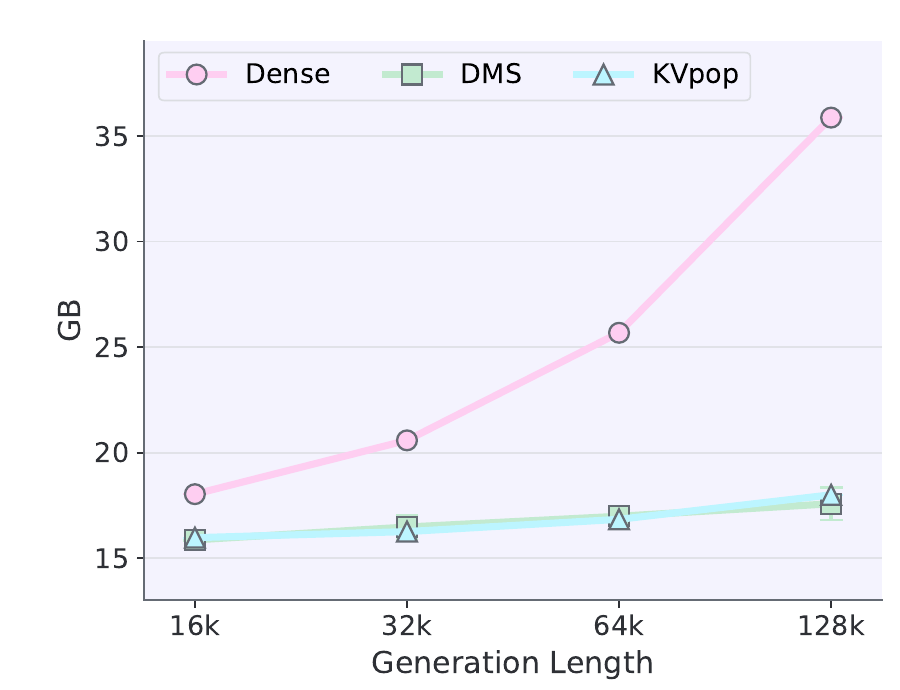}
        \caption{Peak allocated VRAM.}
        \label{fig:inference_vram}
    \end{subfigure}

    \caption{\textbf{Inference efficiency on \QwenThreeEightB.} 
    We measure end-to-end decoding latency and peak allocated VRAM at batch size 1, 75\% \acs{kv} compression, and varying generation lengths. Dense attention incurs growing memory cost as the \acs{kv} cache expands, whereas DMS and \ourmethod~grow more slowly. \ourmethod~further reduces latency, maintaining stable throughput across long generations and outperforming both dense attention and DMS at the longest sequence lengths.}
    \label{fig:inference_overview}
\end{figure*}

\textbf{Inference efficiency.}
We benchmark end-to-end autoregressive decoding at batch size 1 and generation lengths up to 131k tokens, reporting peak allocated VRAM and latency averaged over five steady-state runs (excluding warmup and compilation) in Figure \ref{fig:inference_overview}. For baselines, we use the provided Hugging Face implementations.\footnote{For DMS we use the cache implementation from \href{https://huggingface.co/nvidia/Qwen3-8B-DMS-8x}{https://huggingface.co/nvidia/Qwen3-8B-DMS-8x}.\\As mentioned in Section \ref{sec:experiments} we match parameters of the scorer with \ourmethod.}
Dense attention exhibits the expected linear \acs{kv} growth, with peak VRAM rising from 18GB at 16k tokens to 36GB at 131k, while both DMS and \ourmethod~only grow by 19\% to 19GB.

\enlargethispage{32pt}

\begin{nxaiinfo}[Why \ourmethod~decodes faster than DMS]
\ourmethod~enforces the same fixed budget on \emph{every} \acs{kv} head while DMS lets dynamic gates concentrate capacity in a few heads, yielding ragged per-head caches that are harder to execute and compile (see Figure \ref{fig:dms_sparsity}).
For this reason \ourmethod~sustains higher long-generation throughput at a comparable cache budget.
Evaluating whether this advantage persists under paged \acs{kv}-cache managers such as vLLM~\citep{kwon2023efficient} or SGLang~\citep{zheng2024sglang} remains future work.
\end{nxaiinfo}

\vspace{-4pt}
\subsection{Delayed scoring improves stateful eviction}

\begin{wrapfigure}{r}{0.48\textwidth}
    \vspace{-20.5pt}
    \centering
    \includegraphics[width=0.46\textwidth]{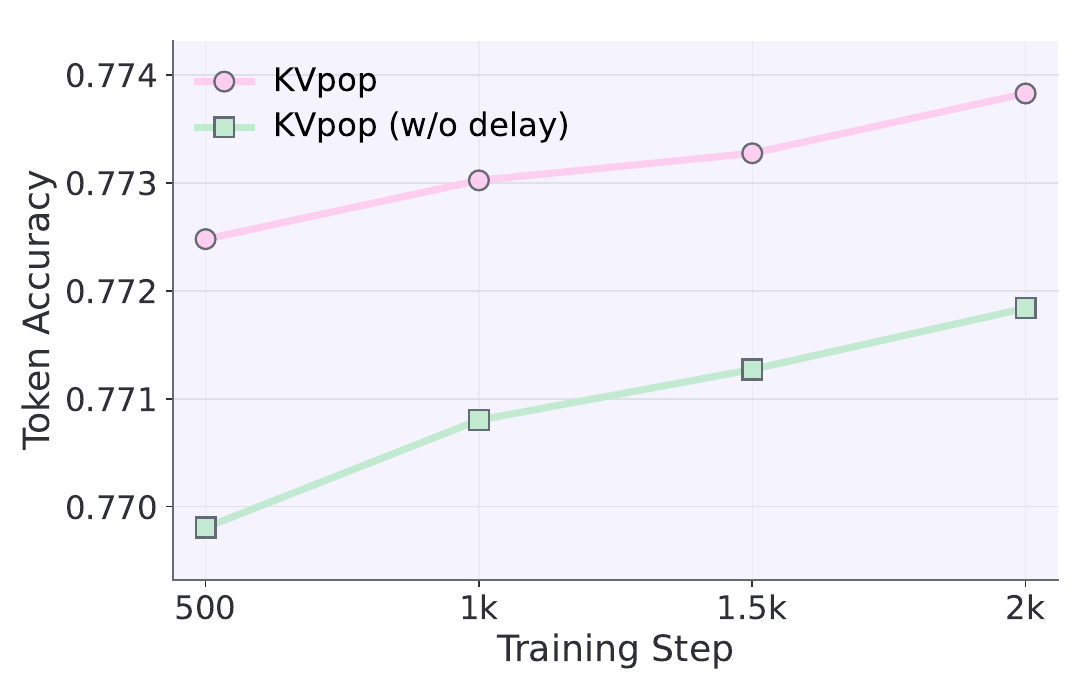}
    \captionsetup{width=0.42\textwidth}
    \setlength{\abovecaptionskip}{2pt}
    \caption{\textbf{Effect of delayed scoring.}
    Token accuracy with and without delayed scoring.}
    \label{fig:arch_ablation}
    \vspace{-16pt}
\end{wrapfigure}

We ablate the effect of delayed scoring in Figure~\ref{fig:arch_ablation}, comparing an \mlstm~scorer with and without delayed readout.
After 2{,}000 steps, delayed \mlstm~scoring achieves a 0.2-point increase in token accuracy over immediate scoring.

A memory-based scorer is beneficial only when its state can integrate additional context before the eviction decision. Without delayed scoring, the \mlstm, like a stateless scorer, must commit before near-future evidence is available. Delayed readout addresses this by aligning the eviction decision with the moment near-future context becomes available.

\subsection{Eviction Policy Analysis}

\begin{figure*}
    \centering
    \includegraphics[width=1.0\linewidth]{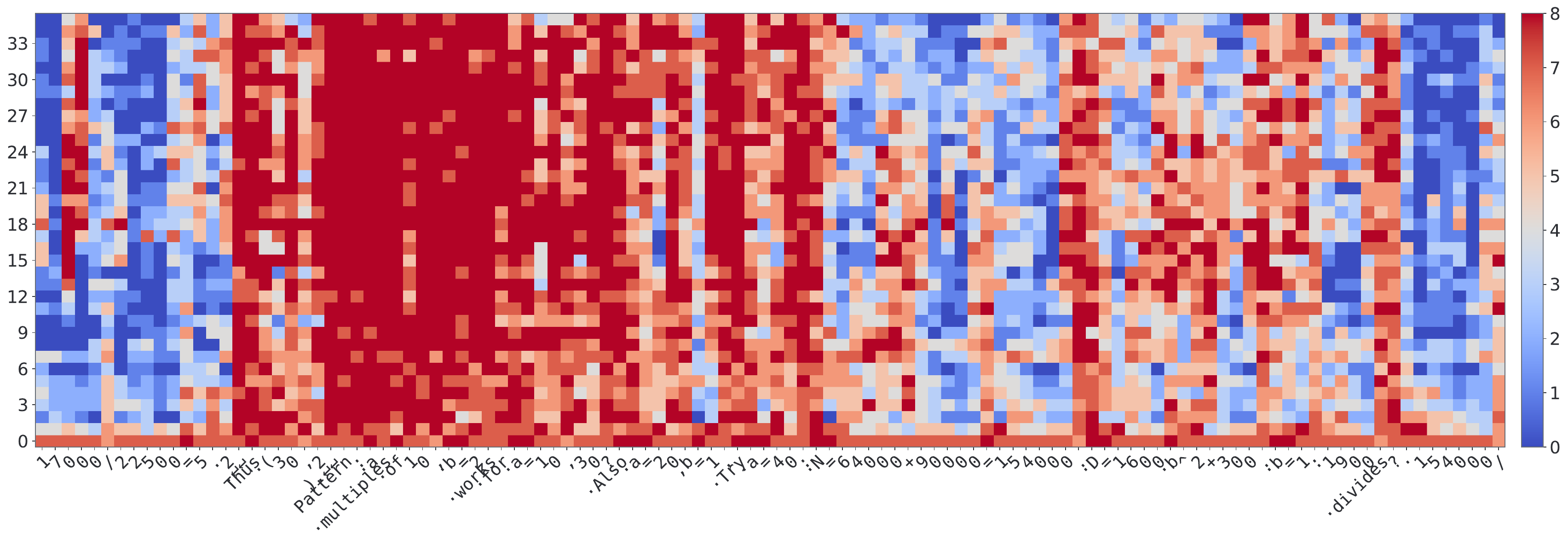}
    \vskip -.1in
    \caption{\textbf{Eviction patterns in a mathematical reasoning trace.}
    We visualize the last 112 tokens of a randomly sampled sequence. Rows denote layers, columns denote token strings, and color indicates how many attention heads in a layer retained the corresponding token. \ourmethod~tends to evict purely numeric tokens more often, while retaining reasoning-structural tokens.}
    \label{fig:evict_pattern}
\end{figure*}

\textbf{Eviction Pattern Visualization.}
Figure~\ref{fig:evict_pattern} visualizes token retention over the final 112 evictable tokens of a randomly sampled mathematical reasoning sequence.
Rows are transformer layers, columns are token strings, and color encodes how many attention heads retain each token.
\ourmethod~does not retain tokens uniformly.
Purely numeric tokens are dropped more often, while tokens that organize or advance the reasoning, such as discourse markers (\textit{Thus}), operation words (\textit{multiplies}), and symbolic tokens (\textit{=}), are retained by more heads and across more layers.
The first layer is a notable exception, retaining nearly all tokens across heads, consistent with prior observations that early layers perform broader processing before higher layers specialize~\citep{tenney2019bert}.
This indicates that \ourmethod~learns a policy beyond simple recency, preferentially keeping tokens useful for interpreting the evolving solution. Additional examples are in Appendix Figure~\ref{fig:eviction_patterns}.
As a complementary comparison, Appendix~\ref{app:dms_sparsity} analyzes the sparsity patterns learned by DMS, showing that its dynamic gates concentrate long-range capacity in a small subset of heads.

\textbf{Recovering oracle eviction decisions.}
In Appendix Figure~\ref{fig:teacher_alignment} we plot the top-$k$ retention recall distribution over heads across all layers of \QwenThreeFourB~at 75\% \acs{kv} cache compression.
For a given head the metric measures what fraction of the full-attention teacher's top-$k$ retained tokens are also kept by the learned \ourmethod~scorer under the same cache budget.
Most layer-wise distributions remain in a high range, with a global mean recall of 81\%, indicating high agreement with the teacher policy.

\section{Conclusion}

\textbf{Limitations.}
While \mlstm~scoring provides a strong stateful \acs{kv} pruning mechanism, we leave a broader exploration of alternative memory-based scorers for future work.
\ourmethod~is designed as an efficient post-training retrofit for Transformers with dense attention, rather than a compressed-cache architecture trained from scratch.
Finally, our homogeneous per-head cache budget enables efficient GPU execution, but hybrid dense-sparse layers may further improve the quality-efficiency trade-off.

\textbf{Conclusion.}
We introduced \ourmethod, a fixed-budget \acs{kv} cache compression method that learns predictive online decisions from future-attention supervision.
By training scorers at the eviction boundary and optionally delaying stateful scoring until near-future context is available, \ourmethod~directly targets the token-retention decision that determines cache membership.
The resulting sparse retrofit bounds inference memory and improves the quality-efficiency trade-off over prior eviction methods.
Across AIME and HMMT, \ourmethod~retains 95\% of dense-attention performance on \QwenThreeFourB~at 75\% \acs{kv} cache compression and 94\% at 88\% compression. On \QwenThreeEightB, retention reaches 95\% and 99\% at the same compression ratios.
Moreover, despite being distilled on mathematical reasoning data, the learned eviction policy remains competitive on code generation and STEM reasoning benchmarks.
These results show that predictive learned eviction can preserve long-context reasoning quality under aggressive cache compression.

\section*{Acknowledgments}
We acknowledge EuroHPC Joint Undertaking for awarding us access to Leonardo at CINECA, Italy, Deucalion at MACC, Portugal, Discoverer at SofiaTech, Bulgaria.
This work was supported by European Union’s Horizon Europe research and innovation programme under grant agreement number 101214398 (ELLIOT).

\printbibliography[heading=nxairefs]

\clearpage
\appendix
\startcontents[appendix]
\section*{Appendix}
\printcontents[appendix]{l}{1}{\setcounter{tocdepth}{2}}
\clearpage

\section{Background: Linear Attention and \mlstm}
\label{app:xlstm}

\textbf{Linear attention} replaces the softmax similarity
$\kappa_{\exp}(\boldsymbol{q},\boldsymbol{k})=\exp\!\big(\boldsymbol{q}^\top \boldsymbol{k}/\sqrt{d_{qk}}\big)$
with a kernel $\kappa_{\phi}(\boldsymbol{q},\boldsymbol{k})=\phi(\boldsymbol{q})^\top \phi(\boldsymbol{k})$ that admits an explicit feature representation, where $\phi:\mathbb{R}^{d_{qk}}\!\to\!\mathbb{R}^{d_{qk}}$ \citep{katharopoulos2020transformers}.

Through associativity, this factorization yields two mathematically equivalent causal attention implementations:
(i) a (chunkwise) parallel computation suited for training and prefill, and
(ii) an online recurrent update used for step-by-step decoding.
By switching between these views, one obtains linear-time prefill and training and constant-memory autoregressive generation \citep{yang2024gated}.

In the recurrent formulation, each head maintains a running key--value summary
$\boldsymbol{C}_t\in\mathbb{R}^{d_{qk}\times d_v}$, optionally together with a normalizer
$\boldsymbol{z}_t\in\mathbb{R}^{d_{qk}}$.
Given the token at time $t$, the state is updated via rank-one outer products:
\begin{align*}
    \boldsymbol{C}_{t} &= \boldsymbol{C}_{t-1}+\phi(\boldsymbol{k}_{t}) \otimes \boldsymbol{v}_{t} \\
    \boldsymbol{z}_{t} &= \boldsymbol{z}_{t-1}+\phi(\boldsymbol{k}_{t})
\end{align*}
where $\otimes$ denotes the outer product.
For a query $\boldsymbol{q}_t$, the head output is obtained by reading from the current summary with normalization:
\begin{equation*}
    \boldsymbol{h}_{t} = \frac{\phi(\boldsymbol{q}_{t}) \boldsymbol{C}_{t}}{\phi(\boldsymbol{q}_{t})\boldsymbol{z}_{t}}
\end{equation*}
Here $\boldsymbol{q}_t,\boldsymbol{k}_t\in\mathbb{R}^{d_{qk}}$ and $\boldsymbol{v}_t\in\mathbb{R}^{d_v}$.

\textbf{\mlstm}. Recently, modern recurrent architectures, such as \xlstm~ \citep{beck2024xlstm}, Mamba \citep{gu2024mamba}, and Gated Delta Networks \citep{yang2024delta} have emerged as competitive linear-complexity alternatives to Transformers. Inspired by the gating structure of the original LSTM cell \citep{hochreiter1997lstm}, these operators augment the outer product update of linear attention with expressive gates.
In this work, we choose \mlstm, which was introduced as a sublayer of \xlstm, as a stateful importance scorer. \mlstm~ has been shown to perform well in language modelling \citep{beck2025xlstm}, computer vision \citep{alkin2024vision}, biological modeling
\cite{schmidinger2024bio}, decision-making \citep{schmied2024large}, and time series forecasting \citep{auer2025tirex}. The \mlstm~ introduces three input-dependent gates into the state update of linear attention, each governing a different part of
the computation. Let $\boldsymbol{w}_i \in \mathbb{R}^{d \times 1}$ and $\boldsymbol{w}_f \in \mathbb{R}^{d \times 1}$ parameterize
scalar input and forget gates, and let $\boldsymbol{W}_{og} \in \mathbb{R}^{d \times d_v}$ parameterize a vector-valued output gate.
Given a token representation $\boldsymbol{x}_t$, we compute
\[
i_t = \exp(\boldsymbol{x}_t \boldsymbol{w}_i), \qquad
f_t = \sigma(\boldsymbol{x}_t \boldsymbol{w}_f), \qquad
\boldsymbol{o}_t = \sigma(\boldsymbol{x}_t \boldsymbol{W}_{og}),
\]
where $i_t$ scales the new key--value write, $f_t$ attenuates the running state, and $\boldsymbol{o}_t$ gates the readout.
The resulting recurrent updates are
\begin{align*}
    \boldsymbol{C}_t &= f_t\,\boldsymbol{C}_{t-1} + i_t\,\phi(\boldsymbol{k}_t)\otimes \boldsymbol{v}_t \\
    \boldsymbol{z}_t &= f_t\,\boldsymbol{z}_{t-1} + i_t\,\phi(\boldsymbol{k}_t)
\end{align*}
with numerical stabilization for the exponential input gate omitted here for clarity.
A query then performs the usual normalized retrieval, and the output gate modulates it elementwise:
\begin{equation}
    \boldsymbol{h}_t
    = \boldsymbol{o}_t \odot
    \frac{\phi(\boldsymbol{q}_{t}) \boldsymbol{C}_{t}}{\phi(\boldsymbol{q}_{t})\boldsymbol{z}_{t}}
    \label{eq:mlstm-read}
\end{equation}

\textbf{Lightweight stateful scoring with \mlstm}. We tailor the \mlstm~scorer for minimal retrofit overhead in a pretrained Transformer.
We remove the \mlstm~output gate, reuse the Transformer’s existing attention projections, and build scorer inputs solely from cached quantities. For each KV head, we form
$x_t=[k_t;v_t]\in\mathbb{R}^{2d_{qkv}}$ and apply a small head-specific linear projection $W_{q/k/v}\in\mathbb{R}^{(2d_{qkv})\times(d_{qkv}/2)}$ followed by a \emph{Hedgehog} activation $\phi$ in its softmax-stabilized form \citep{zhang2024hedgehog}:
\[
\phi(\textbf{x}_t)=\big[\mathrm{softmax}(\textbf{x}_t);\mathrm{softmax}(-\textbf{x}_t)\big]\in\mathbb{R}^{d_{qkv}}
\]
where the softmax is taken over the feature dimension.
Crucially, this adds only one small matrix for each query, key, and value head. We base these \mlstm~ adaptions on \cite{hauzenberger2026xlstmdistillation}.

\section{Target Variants and Boundary-Loss Details}
\label{app:target_loss_details}

This appendix provides details omitted from Section~\ref{sec:target_loss}: GQA
aggregation variants, normalization of the future-attention target, temporal
score decay, and the optional weighting and sampling choices used by the
boundary-aware loss.

\subsection{Future-Attention Target Variants}
\label{app:target_variants}

For each KV head $h$, multiple query heads $g=1,\ldots,G$ attend to the
same key and value vectors. We first compute a per-group future-attention mass,
\begin{equation}
\bar m^{(h,g)}_t
=
\sum_{d=t+w}^{S-1} p^{(h,g)}_{d\to t}
\label{eq:group_future_mass_app}
\end{equation}
which is related to the normalized mass $m^{(h,g)}_t$ from
Eq.~\eqref{eq:future_mass_per_group} by $m^{(h,g)}_t = \bar m^{(h,g)}_t / N_t$.
Because a single KV entry is shared by all $G$ query heads, these per-group
masses must be aggregated into one target score per KV head and token. Our
default choice is max aggregation,
\begin{equation}
\bar m_{h,t}
=
\max_{g} \bar m^{(h,g)}_t
\qquad
\bar r^{\mathrm{tgt}}_{h,t}=\log(\bar m_{h,t}+\epsilon)
\label{eq:max_group_aggregation_app}
\end{equation}
which implements an existential criterion: a token should be retained if any
query head that shares the KV entry strongly relies on it. This is the variant
used in our main experiments unless specified otherwise.

As an alternative, we also consider probability-space mean aggregation,
\begin{equation}
\bar m_{h,t}
=
\frac{1}{G}\sum_{g=1}^{G} \bar m^{(h,g)}_t
\qquad
\bar r^{\mathrm{tgt}}_{h,t}=\log(\bar m_{h,t}+\epsilon)
\label{eq:mean_group_aggregation_app}
\end{equation}
which rewards tokens that are broadly useful across the query heads sharing a KV
head. Mean aggregation can be preferable when average shared utility is more
important than preserving rare head-specific dependencies, but it may underweight
tokens that are crucial for only one query head.

\subsection{Count Normalization}
\label{app:count_normalization}

The unnormalized future-attention mass in Eq.~\eqref{eq:group_future_mass_app}
can be larger for early tokens because they have more future query positions.
When desired, we normalize by the number of future queries that can attend to a
key after it leaves the protected window. For token $t$, this count is
\begin{equation}
N_t = \max(1, S-(t+w))
\label{eq:future_count_app}
\end{equation}
The normalized log-target is therefore
\begin{equation}
r^{\mathrm{tgt}}_{h,t}
=
\bar r^{\mathrm{tgt}}_{h,t} - \log N_t
\label{eq:normalized_target_app}
\end{equation}
This variant measures average rather than total future attention. If a finite
lookahead window $L$ is used in the target computation, the count is clipped to
$N_t \le L+1$.

\subsection{Temporal Decay and Static Priorities}
\label{app:decay_static_priority}

Both the teacher and learned retention policies rank eligible tokens with an
effective score
\begin{equation}
r_{h,t}(q)
=
r_{h,t}
+
\left\lfloor \frac{q-t}{n} \right\rfloor \log\gamma_h,
\qquad
\gamma_h\in(0,1)
\label{eq:effective_score_app}
\end{equation}
where $n$ is the decay step size. For $n=1$, the query-dependent term decomposes
as
\begin{equation}
r_{h,t}(q)
=
\left(r_{h,t}-t\log\gamma_h\right)
+
q\log\gamma_h
\label{eq:static_priority_decomposition_app}
\end{equation}
At fixed $q$, the term $q\log\gamma_h$ is common to all eligible tokens, so the
top-$k$ ranking is equivalent to sorting by the static priority
\begin{equation}
\bar r_{h,t} = r_{h,t}-t\log\gamma_h
\label{eq:static_priority_app}
\end{equation}
For $n>1$, the same idea applies with piecewise-constant age buckets. In our
implementation, the decay factor is learned per KV head by mapping an
unconstrained parameter into a fixed log-decay range,
\begin{equation}
\log\gamma_h
=
\log\gamma_{\min}
+
\sigma(\alpha_h)
\left(\log\gamma_{\max}-\log\gamma_{\min}\right)
\label{eq:decay_param_app}
\end{equation}
with $0<\gamma_{\min}<\gamma_{\max}<1$. This constraint stabilizes training and
prevents degenerate decay values.

\subsection{Margin Weighting}
\label{app:margin_weighting}

Some boundary decisions are ambiguous because the newly eligible token and the
teacher boundary token have nearly equal target scores. We optionally downweight
such examples using the teacher margin
\begin{equation}
\Delta^{\mathrm{tgt}}_{q,h}
=
y_{q,h}
\left(
 r^{\mathrm{tgt}}_{h,t_{\mathrm{new}}}(q)
 -
 r^{\mathrm{tgt}}_{h,t_{\mathrm{bnd}}}(q)
\right)
\ge 0
\label{eq:teacher_margin_app}
\end{equation}
The margin weight is
\begin{equation}
\omega^{\mathrm{margin}}_{q,h}
=
\omega_{\min}
+
(1-\omega_{\min})
\sigma\left(\frac{\Delta^{\mathrm{tgt}}_{q,h}}{\tau_\omega}\right)
\label{eq:margin_weight_app}
\end{equation}
where $\omega_{\min}$ is the minimum example weight and $\tau_\omega$ controls
the sharpness. This keeps high-margin decisions near weight one while reducing
the influence of near-ties.

\subsection{Keep/Drop Balancing}
\label{app:keep_drop_balancing}

The boundary label distribution may be imbalanced for a given head, especially
when the scorer or decay strongly favors either old or newly eligible tokens. We
optionally apply a headwise balancing factor so that keep and drop decisions have
similar aggregate weight. Let $\rho_h$ be the fraction of valid sampled boundary
decisions with $y_{q,h}=+1$ for head $h$. We use
\begin{equation}
\omega^{\mathrm{bal}}_{q,h}
=
\begin{cases}
\operatorname{clip}\left(\frac{1}{2\rho_h}, c_{\min}, c_{\max}\right),
& y_{q,h}=+1 \\
\operatorname{clip}\left(\frac{1}{2(1-\rho_h)}, c_{\min}, c_{\max}\right),
& y_{q,h}=-1
\end{cases}
\label{eq:balance_weight_app}
\end{equation}
and normalize the resulting weights by their mean over valid examples. The final
weight in Eq.~\eqref{eq:boundary_loss} is the product of the margin and balancing
weights.

\subsection{Sampling Query Positions}
\label{app:q_sampling}

Computing the boundary loss for every query position is unnecessary. We sample a
small set of positions after the long-range budget is saturated:
\begin{equation}
q \ge s+w+k .
\label{eq:q_sampling_start_app}
\end{equation}
Our default sampler mixes uniform samples with samples biased toward later
positions. This improves coverage of both early saturation behavior and the
long-range regime where cache pressure is strongest. The teacher ranks and
cutoffs are computed for all positions once, so evaluating the loss at sampled
positions costs only a gather and a pairwise logistic term.

\section{Efficient Running Top-$k$ Sparse Attention}
\label{app:running_topk}

This appendix describes the fixed-budget sparse mask used during training and
prefill. During autoregressive decoding, the cache itself contains only retained
entries, so attention can be computed directly over the compact cache. During
training, however, we need a parallel sparse-attention mask over the full
sequence.

\subsection{Mask Definition}
\label{app:mask_definition}

For each query position $q$, the retained keys are the union of three sets:
\begin{enumerate}
\item sink tokens $\{t:t<s\}$
\item recent-window tokens $\{t:q-t<w\}$
\item the top-$k$ tokens in $\mathcal{E}(q)=\{t:s\le t\le q-w\}$ under the
      predicted effective score
\end{enumerate}
Let $\operatorname{rank}(t)$ be the rank of token $t$ under the static priority
from Eq.~\eqref{eq:static_priority_app}, with smaller ranks indicating higher
priority. Let $\tau_q$ be the cutoff rank of the $k$-th best eligible token at
query $q$. The sparse mask is
\begin{equation}
M_{q,t}
=
\mathbf{1}[t<s]
\lor
\mathbf{1}[q-t<w]
\lor
\left(
\mathbf{1}[t\in\mathcal{E}(q)]
\land
\mathbf{1}[\operatorname{rank}(t)\le \tau_q]
\right)
\label{eq:running_topk_mask_app}
\end{equation}
This mask enforces the fixed budget $B=s+w+k$ once the sequence is longer than
$s+w+k$.

\newpage

\subsection{Fenwick-Tree Cutoff Computation}
\label{app:fenwick_cutoff}

\begin{wrapfigure}{r}{0.52\textwidth}
\vspace{-1.3em}
\hfill
\begin{minipage}{0.50\textwidth}
\centering
\begin{algobox}
\footnotesize
\begin{algorithmic}[1]
\Require
\Statex \hspace{\algorithmicindent} static ranks $\operatorname{rank}(0),\ldots,\operatorname{rank}(S-1)$
\Statex \hspace{\algorithmicindent} budget $(s,w,k)$

\State initialize Fenwick tree $T$ with zeros
\State $\tau_q \gets -1$ for all $q$

\For{$q = 0,\ldots,S-1$}
    \State $t \gets q-w$

    \If{$t \ge s$}
        \State insert $\operatorname{rank}(t)$ into $T$
    \EndIf

    \If{$|\mathcal E(q)| \ge k$}
        \State $\tau_q \gets
        \mathrm{FindByPrefixCount}(T,k)$
    \EndIf
\EndFor

\State \Return $\tau_0,\ldots,\tau_{S-1}$
\end{algorithmic}
\end{algobox}
\captionof{algorithm}{Running top-$k$ cutoff computation for one KV head.
A Fenwick tree maintains eligible long-range tokens ordered by their static
rank. For each query position, the cutoff $\tau_q$ is the rank threshold
corresponding to the current top-$k$ eligible tokens.}
\label{alg:fenwick_topk}
\end{minipage}
\vspace{-2.0em}
\end{wrapfigure}

The cutoff sequence $\tau_q$ can be computed in $O(S\log S)$ time and $O(S)$
space per head. First, sort tokens once by static priority and compute
$\operatorname{rank}(t)$ for all $t$. Then scan query positions from left to
right. When $q$ increases by one, exactly one new token $t=q-w$ enters the
eligible set. We insert its rank into a Fenwick tree over ranks. The cutoff
$\tau_q$ is the smallest rank whose prefix count reaches $k$, i.e. the rank of
the $k$-th highest-priority eligible token. Algorithm~\ref{alg:fenwick_topk} summarizes the procedure.

The same thresholds are used both to instantiate the sparse attention mask and
to construct the teacher boundary token for the loss. Our GPU implementation
parallelizes this procedure over batch elements and KV heads. The Fenwick tree
is implemented in Triton; binary lifting retrieves the $k$-th inserted rank
without materializing an $S\times S$ mask.

\subsection{Use with FlexAttention}
\label{app:flexattention_mask}

Given $\operatorname{rank}(t)$ and $\tau_q$, the predicate in
Eq.~\eqref{eq:running_topk_mask_app} can be evaluated inside a sparse attention
kernel. We use \textsc{FlexAttention} with a mask function that checks whether a
key is a sink, belongs to the recent window, or is an eligible long-range token
whose rank is below the query-specific threshold. The mask is generated on the
fly by the kernel, avoiding explicit storage of a dense boolean matrix.

\section{Efficient Future-Target Computation}
\label{app:efficient_target}

This appendix derives the transposed-attention computation used to obtain the
future-attention target in Eq.~\eqref{eq:future_importance_target}.

\subsection{Dense Target Identity}
\label{app:dense_target_identity}

Fix KV head $h$ and query group $g$. Define the dense causal attention logit
\begin{equation}
\ell^{(h,g)}(d,t)
=
\langle \boldsymbol q^{(h,g)}_d, \boldsymbol k^{(h)}_t\rangle/\sqrt{d_k}
\label{eq:dense_logit_app}
\end{equation}
and the causal log-normalizer
\begin{equation}
\mathrm{LSE}^{(h,g)}_d
=
\log \sum_{t'=0}^{d} \exp\left(\ell^{(h,g)}(d,t')\right)
\label{eq:dense_lse_app}
\end{equation}
Then the causal attention probability is
\begin{equation}
p^{(h,g)}_{d\to t}
=
\exp\left(\ell^{(h,g)}(d,t)-\mathrm{LSE}^{(h,g)}_d\right)
\qquad t\le d
\label{eq:causal_prob_app}
\end{equation}
The unnormalized future-attention mass for key $t$ is
\begin{equation}
\bar m^{(h,g)}_t
=
\sum_{d=t+w}^{S-1}
\exp\left(\ell^{(h,g)}(d,t)-\mathrm{LSE}^{(h,g)}_d\right)
\label{eq:future_mass_app}
\end{equation}
Taking logs gives
\begin{equation}
\log \bar m^{(h,g)}_t
=
\operatorname{LSE}_{d:t+w\le d<S}
\left[\ell^{(h,g)}(d,t)-\mathrm{LSE}^{(h,g)}_d\right]
\label{eq:future_mass_lse_app}
\end{equation}
Thus the target is a log-sum-exp over future queries for each fixed key.

\subsection{Transposed Attention}
\label{app:transposed_attention}

Equation~\eqref{eq:future_mass_lse_app} can be evaluated by a second
attention-like pass with swapped query and key roles. Define transposed inputs
\begin{equation}
\boldsymbol q'_t = \boldsymbol k^{(h)}_t
\qquad
\boldsymbol k'_d = \boldsymbol q^{(h,g)}_d
\label{eq:transposed_inputs_app}
\end{equation}
Their dot product gives the original attention logit $\ell^{(h,g)}(d,t)$. We then
apply a score modifier that subtracts the original query normalizer,
\begin{equation}
\ell^{(h,g)}(d,t)
\mapsto
\ell^{(h,g)}(d,t)-\mathrm{LSE}^{(h,g)}_d
\label{eq:transposed_score_modifier_app}
\end{equation}
and a mask enforcing $d\ge t+w$. The auxiliary log-sum-exp returned for
transposed query position $t$ is exactly Eq.~\eqref{eq:future_mass_lse_app}.
Because the value output of this second pass is not used, the value tensor can be
any tensor with a compatible shape.

\subsection{Sparse Normalizer Approximation}
\label{app:sparse_lse_approx}
If Eq.~\eqref{eq:dense_lse_app} is used, the transposed computation gives the
exact dense future-attention target. To reduce training cost, we may reuse the
log-normalizer returned by the sparse attention pass:
\begin{equation}
\widetilde{\mathrm{LSE}}^{(h,g)}_d
=
\log \sum_{t'=0}^{d}
M_{d,t'}\exp\left(\ell^{(h,g)}(d,t')\right)
\label{eq:sparse_lse_app}
\end{equation}
where $M$ is the fixed-budget sparse mask. Replacing
$\mathrm{LSE}^{(h,g)}_d$ with $\widetilde{\mathrm{LSE}}^{(h,g)}_d$ yields an
approximate target. We find this approximation empirically sufficient, since
the top-$k$ retained set captures most of the softmax mass, so
$\widetilde{\mathrm{LSE}}^{(h,g)}_d$ closely tracks $\mathrm{LSE}^{(h,g)}_d$.
This approximation is also practical because it reuses quantities already
produced by the forward sparse-attention call and avoids an additional dense
causal pass. In settings where exact supervision is desired, the dense
normalizer can be computed instead.

\subsection{Implementation Steps}
\label{app:target_implementation_steps}

The target computation proceeds as follows:
\begin{enumerate}
\item Run the main attention pass and obtain per-query log-normalizers. These may
      be dense causal normalizers or sparse normalizers, depending on the chosen
      target variant.
\item Run a transposed attention pass with keys as queries and queries as keys.
\item In the transposed pass, subtract the original per-query log-normalizer as a
      score modifier and mask out positions with $d<t+w$.
\item Use the auxiliary log-sum-exp from the transposed pass to obtain
      $\log \bar m^{(h,g)}_t$ for all keys $t$.
\item Aggregate across query groups according to
      Appendix~\ref{app:target_variants}, and apply count normalization
      (Appendix~\ref{app:count_normalization}) to recover $m^{(h,g)}_t$ as
      defined in the main paper.
\end{enumerate}
The transposed pass is used only while training the scorer. It is not required
for autoregressive decoding.

\section{Scorer Architecture Details}
\label{app:scorer_architecture}

This appendix describes the stateless and stateful scorer variants used with the
same target and boundary loss.

\subsection{Headwise Inputs}
\label{app:headwise_inputs}

For each KV head $h$ and token $t$, we form the scorer input from cached
quantities,
\begin{equation}
\boldsymbol{x}_{h,t}
=
[\boldsymbol{k}_{h,t};\boldsymbol{v}_{h,t}]
\label{eq:scorer_input_app}
\end{equation}
Using $[\boldsymbol{k};\boldsymbol{v}]$ has two advantages. First, it aligns the
scorer with the KV heads used in grouped-query attention. Second, it enables
delayed scoring without storing additional hidden states, because both keys and
values are already present in the cache until the token's eviction decision is
made.

\subsection{Stateless Linear and MLP Scorers}
\label{app:stateless_scorers}

A stateless scorer predicts each token's score independently:
\begin{equation}
\hat r_{h,t}=f^{(h)}_\theta(\boldsymbol{x}_{h,t})
\label{eq:stateless_scorer_app}
\end{equation}
We consider a headwise linear layer and a small headwise MLP. These scorers add
minimal overhead and are easy to parallelize during training and prefill. Their
limitation is that they cannot use context that arrives after token $t$ but
before $t$ becomes evictable.

\subsection{\mlstm~Scorer}
\label{app:mlstm_scorer}

Our memory-based scorer uses an \mlstm-style recurrent state for each KV head.
Given the input $\boldsymbol{x}_{h,t}$, small headwise projections produce
features for the recurrent write and delayed read:
\begin{equation}
\tilde{\boldsymbol q}_{h,t}
=
\phi(W_q^{(h)}\boldsymbol{x}_{h,t})
\qquad
\tilde{\boldsymbol k}_{h,t}
=
\phi(W_k^{(h)}\boldsymbol{x}_{h,t})
\qquad
\tilde{\boldsymbol v}_{h,t}
=
W_v^{(h)}\boldsymbol{x}_{h,t}
\label{eq:mlstm_inputs_app}
\end{equation}
The feature map $\phi$ may be a softmax feature map or the Hedgehog feature map.
For the Hedgehog variant, the projected feature dimension is halved before
applying positive and negative softmax features, reducing parameter overhead.

The recurrent state consists of a matrix memory $\boldsymbol{C}_{h,t}$ and a
normalizer $\boldsymbol{z}_{h,t}$. With scalar input and forget gates $i_{h,t}$
and $f_{h,t}$, the update is
\begin{align}
\boldsymbol{C}_{h,t}
&=
f_{h,t}\boldsymbol{C}_{h,t-1}
+
i_{h,t}\tilde{\boldsymbol k}_{h,t}\otimes \tilde{\boldsymbol v}_{h,t}
\label{eq:mlstm_c_update_app}\\
\boldsymbol{z}_{h,t}
&=
f_{h,t}\boldsymbol{z}_{h,t-1}
+
i_{h,t}\,\tilde{\boldsymbol k}_{h,t}
\label{eq:mlstm_z_update_app}
\end{align}
The gates are produced by headwise projections of $\boldsymbol{x}_{h,t}$:
\begin{equation}
i_{h,t}=\exp(\alpha^{(i)}_{h,t})
\qquad
f_{h,t}=\sigma(\alpha^{(f)}_{h,t})
\label{eq:mlstm_gates_app}
\end{equation}
with optional softcapping of the preactivations for numerical stability. In the
implementation, we use the numerically stabilized \mlstm formulation.

\subsection{Delayed Readout}
\label{app:delayed_readout}

At query position $q$, the newly eligible token is $u=q-w$. The recurrent state
has already processed tokens up to $q$, while the read query is derived from the
newly eligible token $u$:
\begin{equation}
\boldsymbol{h}_{h,u\mid q}
=
\frac{
\tilde{\boldsymbol q}_{h,u}^{\top}\boldsymbol{C}_{h,q}
}{
\tilde{\boldsymbol q}_{h,u}^{\top}\boldsymbol{z}_{h,q}
}
\label{eq:delayed_readout_app}
\end{equation}
The raw score is
\begin{equation}
\hat r_{h,u}
=
\boldsymbol{a}_h^\top
\operatorname{SiLU}(\operatorname{Norm}(\boldsymbol{h}_{h,u\mid q}))
+
b_h
\label{eq:mlstm_output_score_app}
\end{equation}
where the normalization is optional. This time-shifted readout is causal because
all information in $\boldsymbol{C}_{h,q}$ is available at query position $q$.
It is nevertheless forward-looking relative to token $u$, since the state
summarizes the protected-window context observed before $u$ becomes evictable.

\subsection{Initialization and Practical Variants}
\label{app:init_variants}

We initialize the scorer so that the recurrent projections start close to simple
identity-style mappings from the base attention keys and values. The query and
key scorer projections are initialized from the key pathway, the value projection
from the value pathway, and the final score projection can be zero-initialized so
that the sparse policy starts from a neutral ranking before training. Gate biases
are initialized to encourage stable memory updates early in training.

We also evaluate several practical variants: value projection versus direct use
of cached values, optional skip connections from the delayed token to the score
projection, optional headwise normalization, softcapping of gate preactivations,
and stateless MLP scoring. All variants are trained with the same target and
boundary loss, allowing us to isolate the effect of memory-based delayed scoring.

\section{Experiment Details}
\label{app:experiment_details}

\paragraph{Training details.}
All experiments were run on 8 H100 GPUs using PyTorch FSDP.
We used a global batch size of 128 via gradient accumulation, mixed precision (bfloat16 for model parameters and most compute; float32 for gradient all-gather/reductions), and gradient clipping to 1.0 for full finetuning.
For numerical stability, we compute score decay and top-$k$ ranking/selection in float32; all other activations remain in bfloat16.

To maximize GPU utilization, we pack multiple samples into a single sequence up to the maximum context length.
We found that \emph{preserving the attention mask across packed segments} (i.e., not resetting attention at packing boundaries) improved performance for our hybrid architecture, consistent with observations in prior work~\citep{buitrago2025understanding}.

We train \ourmethod{}by minimizing the KL Divergence to the teacher.
For efficiency we compute the KL term over the top-$256$ teacher logits.
This truncation substantially reduces the cost of teacher supervision and suggests a variant where teacher logits could be precomputed offline, avoiding loading the teacher model during training. We leave this optimization for future work.

\begin{table}[t]
\centering
\caption{Training and Inference Hyperparameters.}
\label{tab:hyperparameters}
\begin{tabular}{ll}
\toprule
\multicolumn{2}{l}{\textbf{Training General}} \\
\midrule
Token budget & 2B \\
Context size & $16384$ \\
Batch size & $128$ \\
Weight decay & $0.0001$ \\
Optimizer & AdamW \\
Adam Betas & ($0.9$, $0.95$) \\
Trainable Parameters & all \\
Sequence packing & true \\
Learning rate (existing params) & Constant LR $8\mathrm{e}{-5}$ \\
\makecell[tl]{Learning rate (new params)} &
\makecell[tl]{100 Steps Warmup + Cosine Schedule\\($1\mathrm{e}{-3}$ to $8\mathrm{e}{-5}$)} \\
KL temperature & $1$ \\
KL reverse & false \\
KL Top-$k$ & $256$ \\
\midrule
\multicolumn{2}{l}{\textbf{Training \ourmethod}} \\
\midrule
Decay step size & $1$ \\
Pairwise loss temperature & $1.0$ \\
Pairwise loss margin weighting & true \\
Sliding window & $256$ \\
Sink tokens & $4$ \\
Top-$k$ budget & $2016$ \& $4032$ \\
Decay min & $0.999$ \\
Decay max & $0.999999$ \\
\midrule
{\textbf{Inference}} \\
\midrule
Temperature & $0.6$ \\
Top-$p$ & $0.95$ \\
Top-$k$ & $20$ \\
Min-$p$ & $0.0$ \\
Repetition penalty & $1.0$ \\
Number of shots & $0$ \\
\bottomrule
\end{tabular}

\end{table}

We evaluate all models with the same decoding configuration to ensure that performance differences reflect the retention policy rather than sampling variability.
We use nucleus sampling with temperature $0.6$, top-$p$ $0.95$, top-$k$ $20$, no repetition penalty, and a zero-shot prompt format.

Table \ref{tab:hyperparameters} summarizes the hyperparameters used throughout our training and evaluation pipeline.

\section{\ourmethod{}Sparsity Patterns}
\label{app:eviction_patterns}

\begin{figure*}[p]
    \vspace*{-0.5cm}
    \centering
    \newcommand{\heatmapheight}{0.165\textheight}    \includegraphics[width=\linewidth]{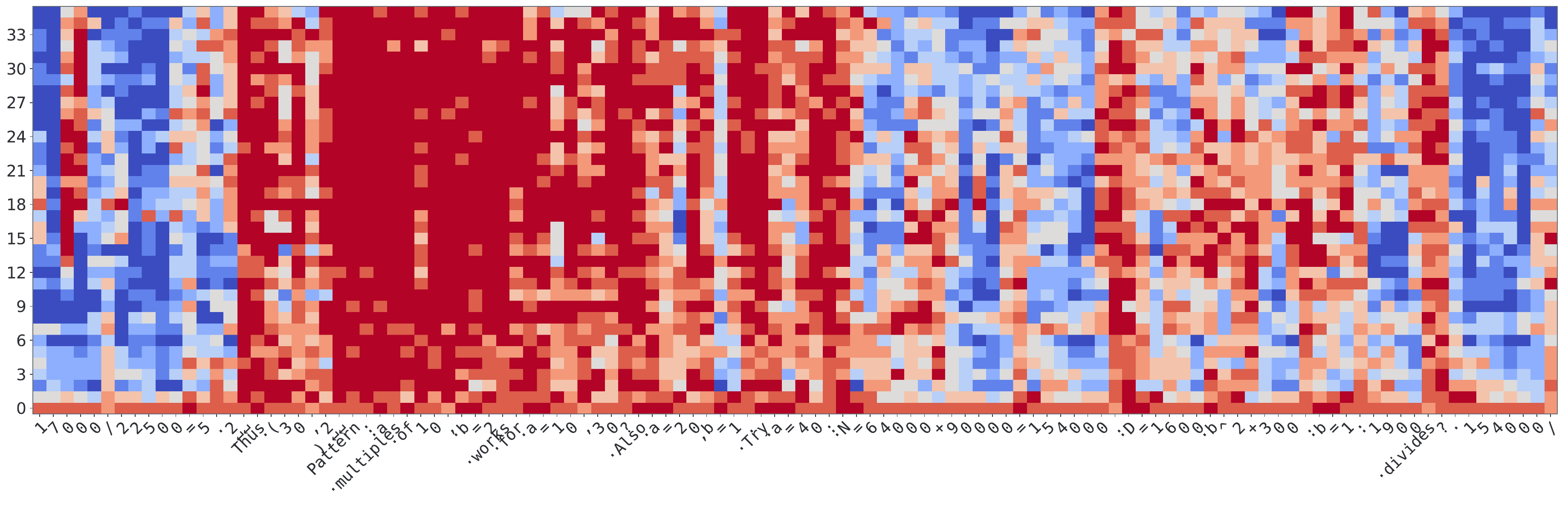}\\[-2mm]    \includegraphics[width=\linewidth]{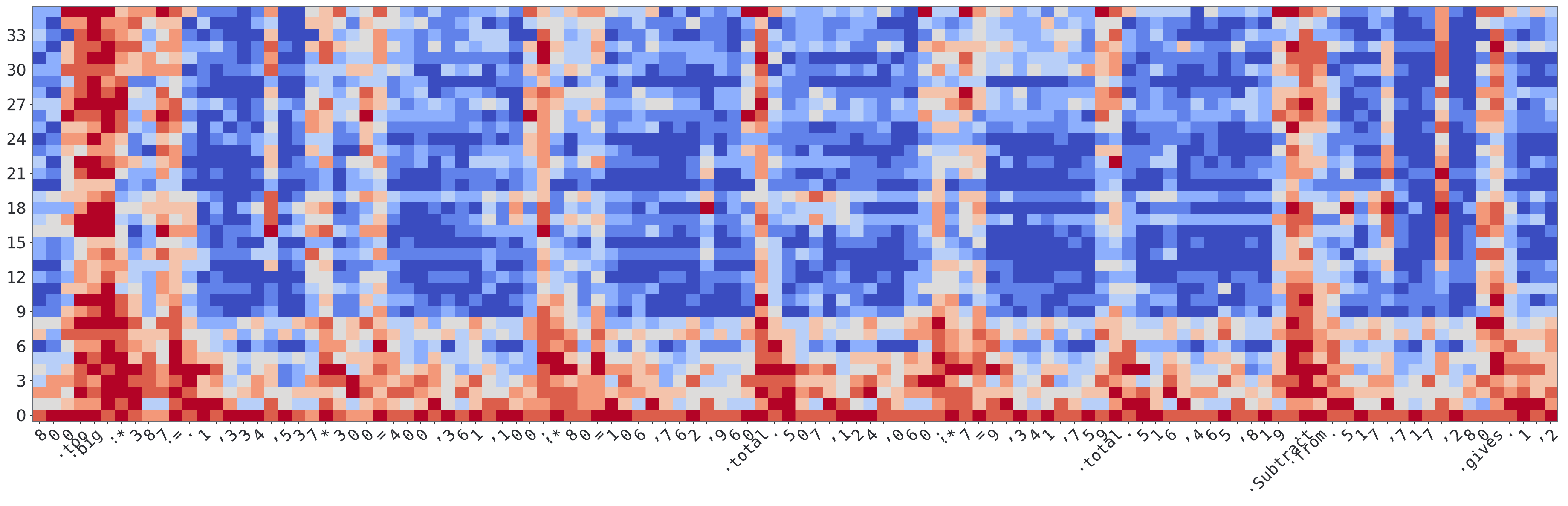}\\[-2mm]    \includegraphics[width=\linewidth]{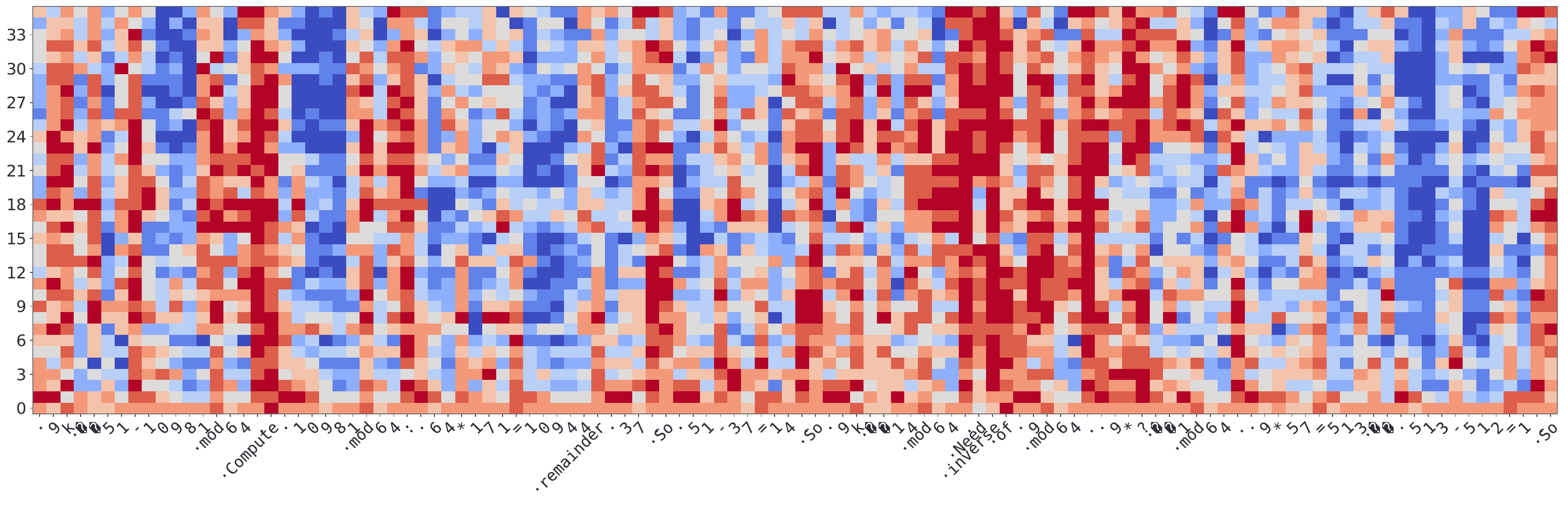}\\[-2mm]    \includegraphics[width=\linewidth]{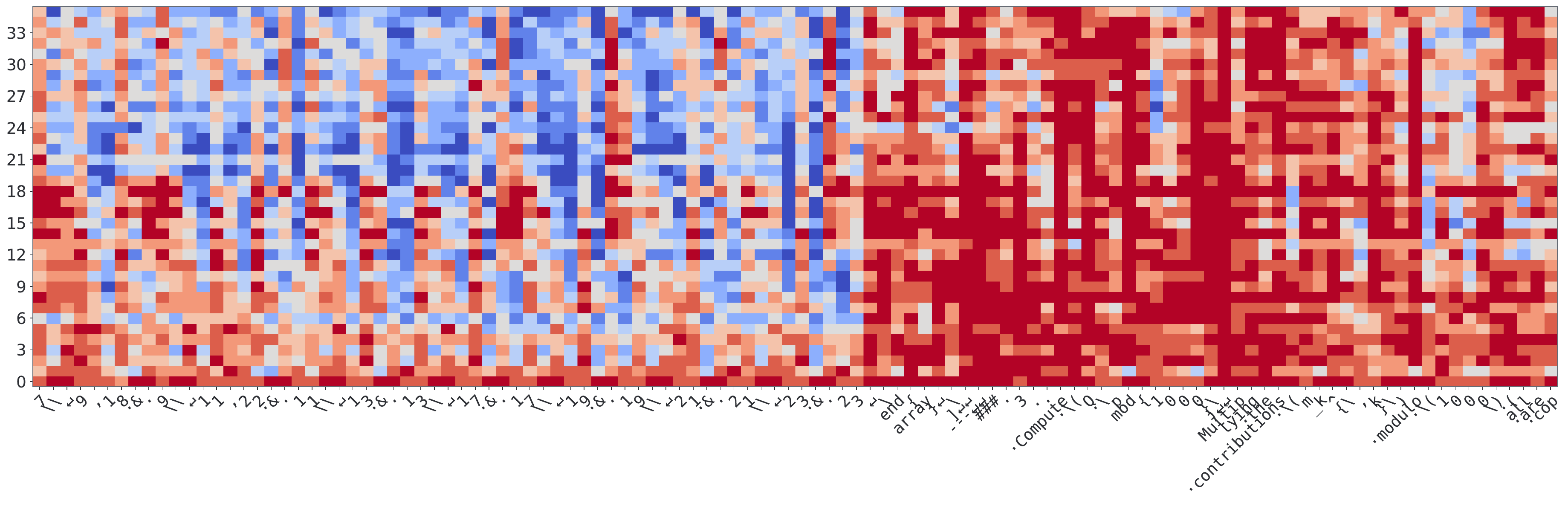}\\[-2mm]    \includegraphics[width=\linewidth]{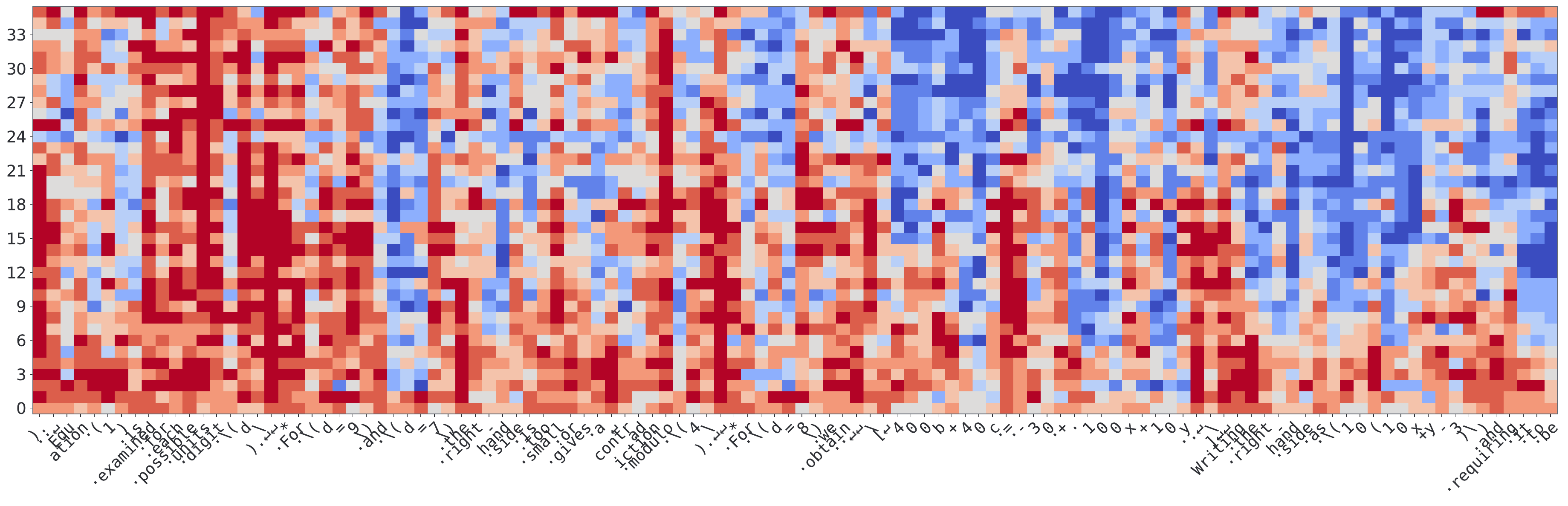}\\[-2mm]
    \includegraphics[width=.65\linewidth,keepaspectratio]{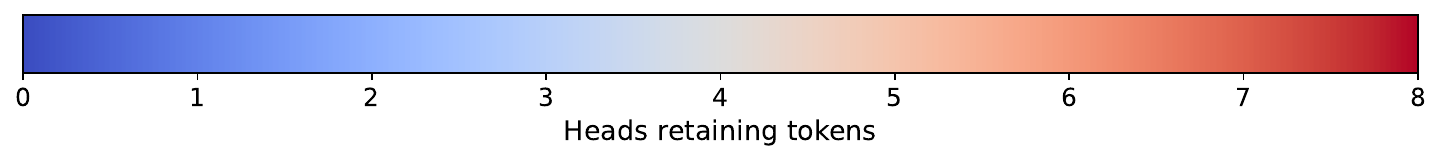}
    \caption{\textbf{\ourmethod{}eviction patterns} (\QwenThreeFourB, CR=75\%, final 112 tokens of five sampled sequences). Color shows how many heads retained each token (x-axis) per layer (y-axis).}
    \label{fig:eviction_patterns}
\end{figure*}

Figure~\ref{fig:eviction_patterns} shows additional qualitative examples of the learned token retention patterns. Across sequences, \ourmethod{}does not simply preserve the most recent tokens or apply a uniform sparsity pattern. Instead, it often retains contiguous blocks of text that appear to carry the structure of the reasoning trace, while evicting many intermediate computation steps and purely numerical tokens once they have served their local role. This behavior is especially visible in later layers, where retained tokens tend to cluster around symbolic expressions, operation words, discourse markers, and other tokens that connect or summarize parts of the derivation. These examples further support the observation that \ourmethod{}learns a content-dependent eviction policy that preferentially preserves tokens useful for interpreting the evolving solution, rather than treating all operands and intermediate values equally.

\begin{figure*}
    \centering
    \includegraphics[width=\linewidth]{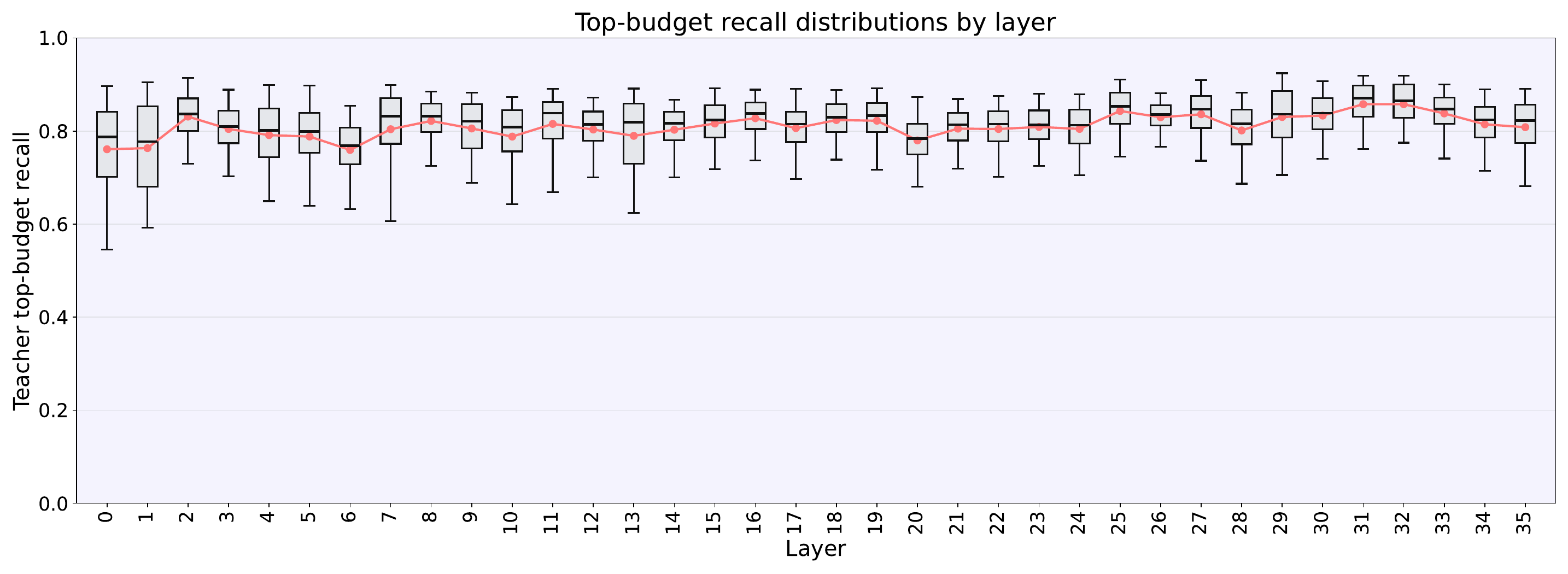}
    \caption{\textbf{Alignment with the future-attention oracle.} We plot the top-$k$ retention recall distribution over heads for all layers of \QwenThreeFourB~at 75\% KV-cache compression. The metric measures what fraction of the full-attention teacher's top-$k$ retained tokens are also kept by the learned \ourmethod{}scorer under the same cache budget. Higher values indicate better agreement with the teacher retention policy.}
    \label{fig:teacher_alignment}
\end{figure*}

\begin{figure*}
    \centering
    \includegraphics[width=1.0\linewidth]{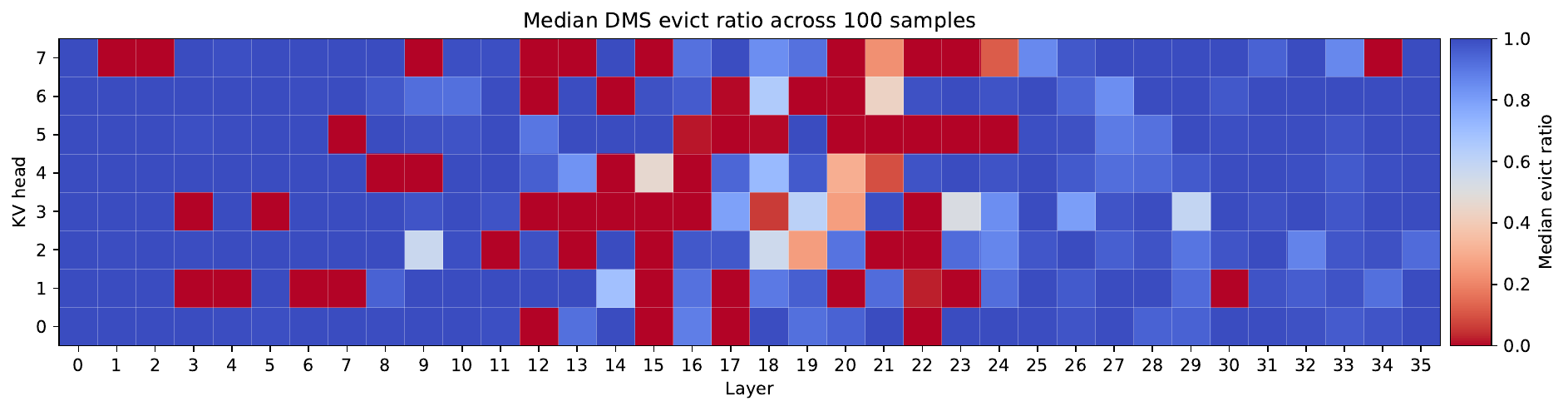}
    \caption{\textbf{DMS evict ratio.} We compute the median eviction ratio over 100 randomly sampled sequences for \QwenThreeFourB~with DMS at a 75\% compression ratio. A value of 0 indicates dense/full attention, whereas a value of 1 indicates that attention is restricted to the local sliding window. DMS exhibits highly heterogeneous eviction patterns: some heads use the full long-range token budget and behave close to dense-attention heads, while many heads—especially in early and late layers—collapse to sliding-window-only attention.}
    \label{fig:dms_sparsity}
\end{figure*}

\section{DMS Sparsity Patterns}
\label{app:dms_sparsity}

Dynamic Memory Sparsification (DMS) learns data-dependent eviction gates and therefore allows the effective cache allocation to vary across heads and layers~\citep{lancucki2025inference}. To understand how a retrofit model uses this dynamic allocation in practice, we train DMS under the same 75\% compression setting as \ourmethod{}and analyze the resulting attention patterns for \QwenThreeFourB. Specifically, we compute the median eviction ratio over 100 randomly sampled sequences, where 0 indicates dense/full attention and 1 indicates that attention is restricted to the local sliding window.

Shown in Figure \ref{fig:dms_sparsity}, DMS exhibits highly heterogeneous eviction patterns. A small subset of heads uses nearly the full long-range token budget and behaves close to dense attention, while many other heads—especially in early and late layers—collapse to sliding-window-only attention. This suggests a winner-takes-all allocation dynamic, where the available long-range budget is concentrated in a few heads rather than distributed evenly across the model. This behavior contrasts with \ourmethod, where each head receives a fixed long-range token allocation by construction, yielding homogeneous and explicitly controlled sparse-attention budget.

\section{\ourmethod{}Pseudocode}
\label{app:kvpop_pseudocode}

The pseudocode summarizes the \ourmethod{}forward pass used for training: each KV entry receives a learned retention score, and sparse attention keeps sinks, the protected recent window, and the highest-ranked long-range entries under a running prefix top-k rule. During training, the same sparse attention pass also provides the log-normalizers used to estimate each key’s future attention mass by a transposed \textsc{FlexAttention} call. This self-targeted signal trains the scorer to preserve keys that continue to receive probability mass after they leave the protected window.

\begin{algorithm}[t]
\caption{\ourmethod{} in PyTorch-style pseudocode.}
\label{alg:kvpop_pseudocode}
\begin{pytorchbox}
def kvpop_attention(Q, K, V, scorer, cfg, training):
    # Predict one score per KV entry. With delayed \mlstm~ scoring,
    # key t is scored when it reaches the eviction boundary.
    r = scorer(K, V, delay=cfg.window)

    # Prefix top-k over decayed KV scores.
    rank, cutoff = running_prefix_topk(
        r,
        sinks=cfg.sinks,
        window=cfg.window,
        topk=cfg.topk,
        decay=cfg.decay,
        decay_step=cfg.decay_step,
    )

    def sparse_mask(b, qh, q, t):
        kh = kv_head(qh)

        sink = t < cfg.sinks
        recent = q - cfg.window < t <= q
        eligible = cfg.sinks <= t <= q - cfg.window
        selected = rank[b, kh, t] <= cutoff[b, kh, q]

        return t <= q and (sink or recent or (eligible and selected))

    # Sparse student attention.
    Y, sparse_lse = flex_attention(
        Q, K, V,
        block_mask=create_block_mask(sparse_mask),
        return_lse=True,
    )

    if not training:
        return Y

    # Reuse the sparse-attention denominators to estimate each key's
    # future attention mass under the current sparse policy.
    def future_mask(b, qh, t, q):
        return q >= t + cfg.window

    def logprob_score(score, b, qh, t, q):
        return score - sparse_lse[b, qh, q]

    _, log_mass = flex_attention(
        repeat_kv_heads(K),      # original keys become queries
        Q,                       # original queries become keys
        zeros_like(Q),
        score_mod=logprob_score,
        block_mask=create_block_mask(future_mask),
        return_lse=True,
    )

    target = log_mass[..., :-cfg.window]
    target = target - log_num_future_queries_per_key(cfg.window)
    target = aggregate_query_heads(target, mode=cfg.group_agg)

    loss = prefix_topk_boundary_loss(r, target, cfg)

    return Y, loss
\end{pytorchbox}
\end{algorithm}

\end{document}